\documentclass[10pt,twocolumn,letterpaper]{article}

\usepackage{cvpr}              %

\usepackage{times}
\usepackage{graphicx}
\usepackage{amsmath}
\usepackage{amssymb}
\usepackage{pifont}
\usepackage{array}
\usepackage{comment} 
\usepackage{capt-of}
\usepackage{booktabs}
\usepackage{tabularx}
\usepackage{soul}
\usepackage{multirow}
\usepackage{multicol}
\usepackage[dvipsnames]{xcolor}
\usepackage{xhfill}

\usepackage[toc,page,header]{appendix}
\usepackage{minitoc}

\usepackage[pagebackref,breaklinks,colorlinks]{hyperref}

\usepackage[capitalize]{cleveref}
\crefname{section}{Sec.}{Secs.}
\Crefname{section}{Section}{Sections}
\Crefname{table}{Table}{Tables}
\crefname{table}{Tab.}{Tabs.}

\makeatletter
\DeclareRobustCommand\onedot{\futurelet\@let@token\@onedot}
\def\@onedot{\ifx\@let@token.\else.\null\fi\xspace}

\makeatother
\let\shortcite\cite

\newcommand{\ap}[1]{``#1''}

\DeclareMathOperator*{\argmin}{arg\,min}
\DeclareMathOperator*{\argmax}{arg\,max}

\begin{document}

\title{Concept Decomposition for Visual Exploration and Inspiration}

\vspace{-1cm}
\author{
Yael Vinker$^{*1,2}$
\and
Andrey Voynov$^{2}$
\and
Daniel Cohen-Or$^{1,2}$
\and
Ariel Shamir$^{3}$
\and \small $^{1}$Tel Aviv University \and  \small$^{2}$Google Research \and  \small$^{3}$Reichman University
\and 
\small\url{https://inspirationtree.github.io/inspirationtree/}
}

\doparttoc %
\faketableofcontents %

\twocolumn[{%
\maketitle

\renewcommand\twocolumn[1][]{#1}%
\begin{center}
    \centering
    \includegraphics[width=0.94\linewidth]{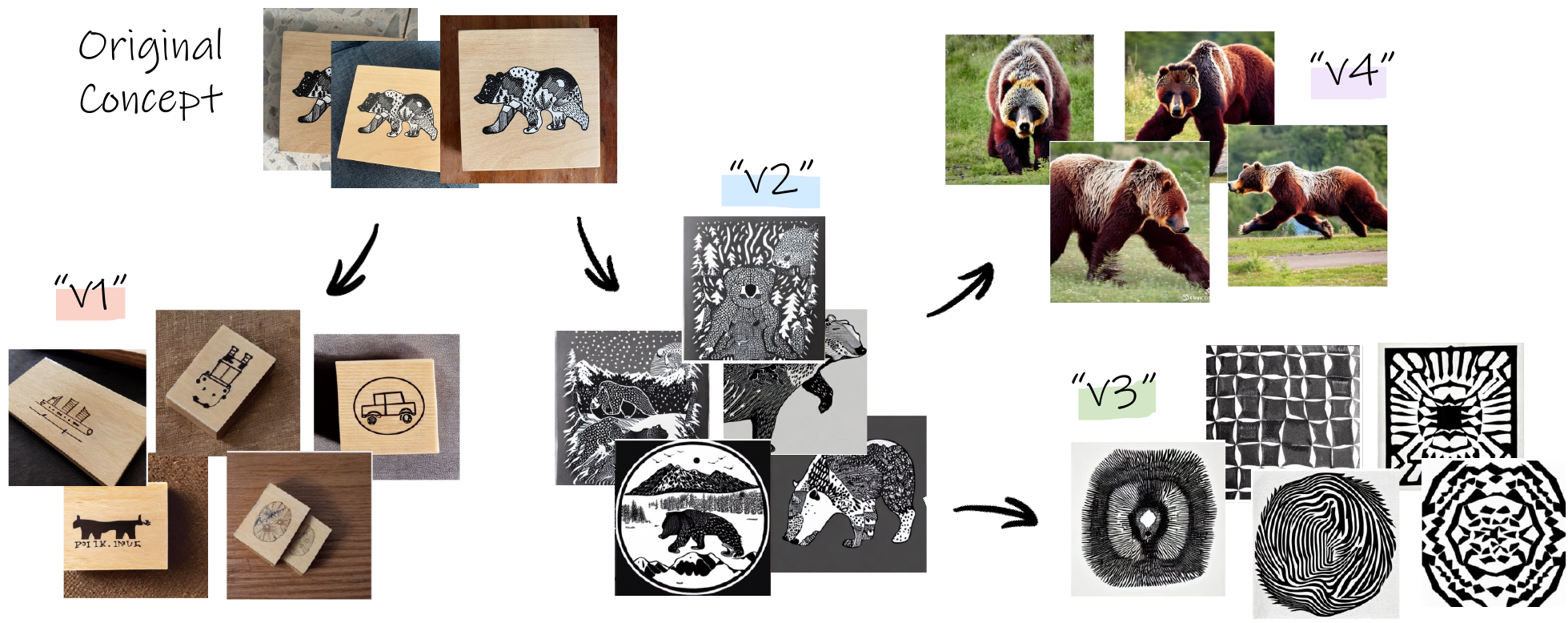}
    \captionsetup{type=figure}\caption{
    Our method provides a tree-structured visual exploration space for a given unique concept. The nodes of the tree (\ap{$v_i$}) are newly learned textual vector embeddings, injected to the latent space of a pretrained text-to-image model.
    The nodes encode different \textit{aspects} of the subject of interest. 
    Through examining combinations within and across trees, the different aspects can inspire the creation of new designs and concepts.
    }
    \label{fig:teaser}
\end{center}%
}]

\def\thefootnote{*}\footnotetext{Work was done during an internship at Google Research}

\thispagestyle{plain}
\pagestyle{plain}

\begin{abstract}
A creative idea is often born from transforming, combining, and modifying ideas from existing visual examples capturing various concepts.
However, one cannot simply copy the concept as a whole, and inspiration is achieved by examining certain aspects of the concept. Hence, it is often necessary to separate a concept into different aspects to provide new perspectives.
In this paper, we propose a method to decompose a visual concept, represented as a set of images, into different visual \textit{aspects} encoded in a hierarchical tree structure. We utilize large vision-language models and their rich latent space for concept decomposition and generation. 
Each node in the tree represents a sub-concept using a learned vector embedding injected into the latent space of a pretrained text-to-image model. We use a set of regularizations to guide the optimization of the embedding vectors encoded in the nodes to follow the hierarchical structure of the tree.
Our method allows to explore and discover new concepts derived from the original one. The tree provides the possibility of endless visual sampling at each node, allowing the user to explore the hidden sub-concepts of the object of interest.
The learned aspects in each node can be combined within and across trees to create new visual ideas, and can be used in natural language sentences to apply such aspects to new designs.
\end{abstract}

\section{Introduction}
\begin{figure}[t]
    \centering
    \includegraphics[width=1\linewidth]{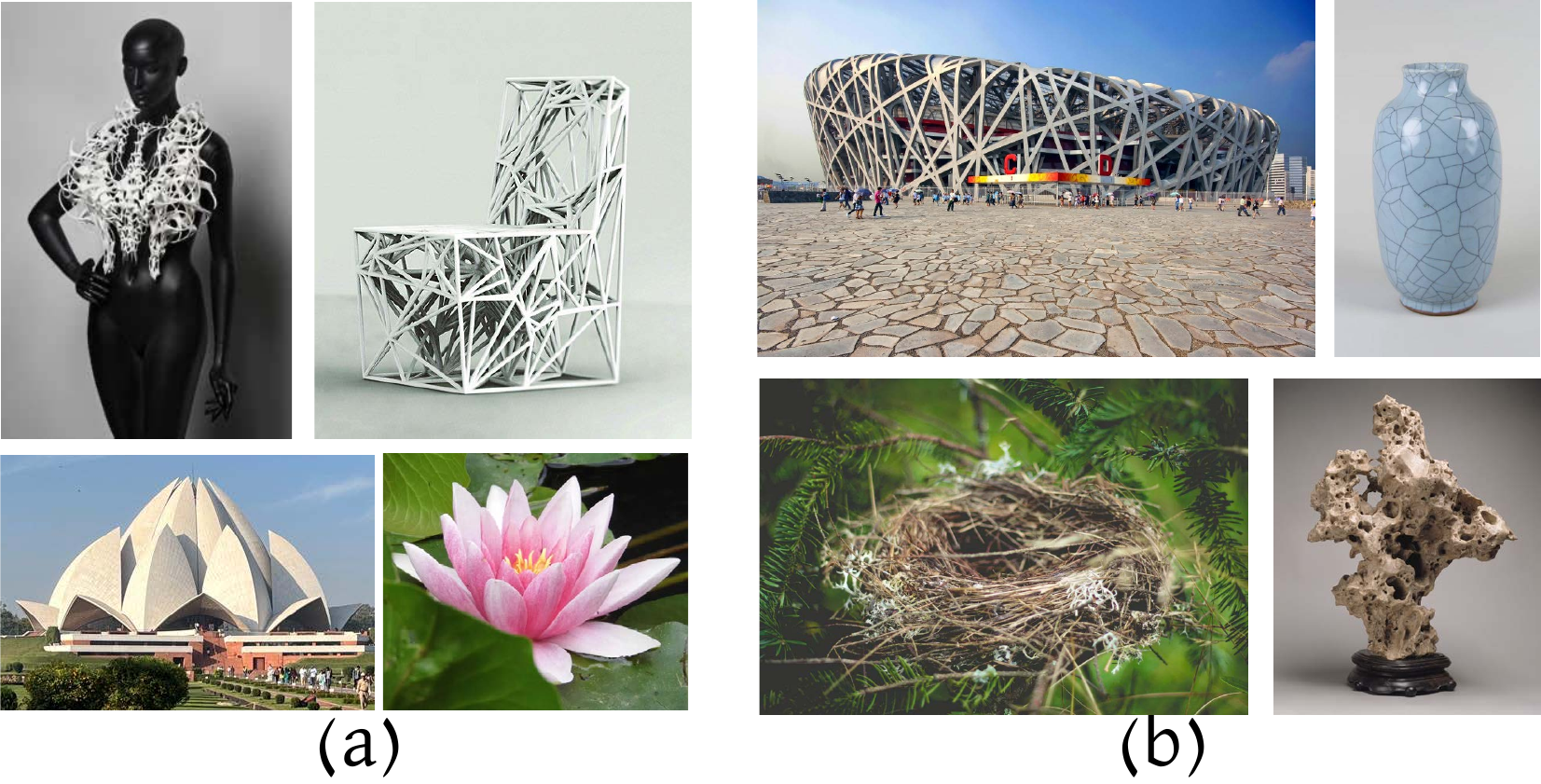}
    \caption{Example of design inspired by visual concepts taken from other concepts.
    (a) top left - fashion design by Iris Van Herpen and Chair by Emmanuel Touraine inspired by nature patterns, bottom left - the Lotus Temple in India, inspired by the lotus flower (b) Beijing National Stadium is inspired by a combination of local Chinese art forms - the crackle glazed pottery that is local to Beijing, and the heavily veined Chinese scholar stones.
    \copyright Dress by Iris van Herpen, chair by Emmanuel Touraine from Wikimedia. Lotus flower, temple, cracked pottery, scholar stone, and bird nest are from rawpixel.com [Public Domain]. Beijing National Stadium photograph by Wojtek Gurak from Flickr.}
    \label{fig:design_inpiration_examples}
\end{figure}

Modeling and design are highly creative processes that often require inspiration and exploration~\cite{Gonalves2014WhatID}. 
Designers often draw inspiration from existing visual examples and concepts - either from the real world or using images \cite{henderson1999line, MULLER198912, ECKERT2000523}.
However, rather than simply replicating previous designs, the ability to extract only certain aspects of a given concept is essential to generate original ideas. For example, in \Cref{fig:design_inpiration_examples}a, we illustrate how designers may draw inspiration from patterns and concepts found in nature. 

Additionally, by combining multiple aspects from various concepts, designers are often able to create something new. 
For instance, it is described \cite{Beijing_National_Stadium} that the famous Beijing National Stadium, also known as the ``Bird's Nest'', was designed by a group of architects that were inspired by various aspects of different Chinese concepts (see \Cref{fig:design_inpiration_examples}b).
The designers combined aspects of these different concepts -- the shape of a nest, porous Chinese scholar stones, and cracks in glazed pottery art that is local to Beijing, to create an innovative architectural design. 
Such a design process is highly exploratory and often unexpected and surprising.

The questions we tackle in this paper is whether a machine can assist humans in such a highly creative process? Can machines understand different aspects of a given concept, and provide inspiration for modeling and design? 
Our work explores the ability of large vision-language models to do just that - express various concepts visually, decompose them into different aspects, and provide almost endless examples that are inspiring and sometimes unexpected. 

We rely on the rich semantic and visual knowledge hidden in large language-vision models.
Recently, these models have been used to perform personalized text-to-image generation \cite{gal2022textual, ruiz2023dreambooth,kumari2022customdiffusion}, demonstrating unprecedented quality of concept editing and variation.
We extend the idea presented in \cite{gal2022textual} to allow \textit{aspect-aware} text-to-image generation, which can be used to visually explore new ideas derived from the original concept.

Our approach involves (1) decomposing a given visual concept into different aspects, creating a hierarchy of sub-concepts, (2) providing numerous image instances of each learned aspect, and (3) allowing to explore combinations of aspects within the concept and across different concepts.

\begin{figure}[b]
    \centering
    \includegraphics[width=0.9\linewidth]{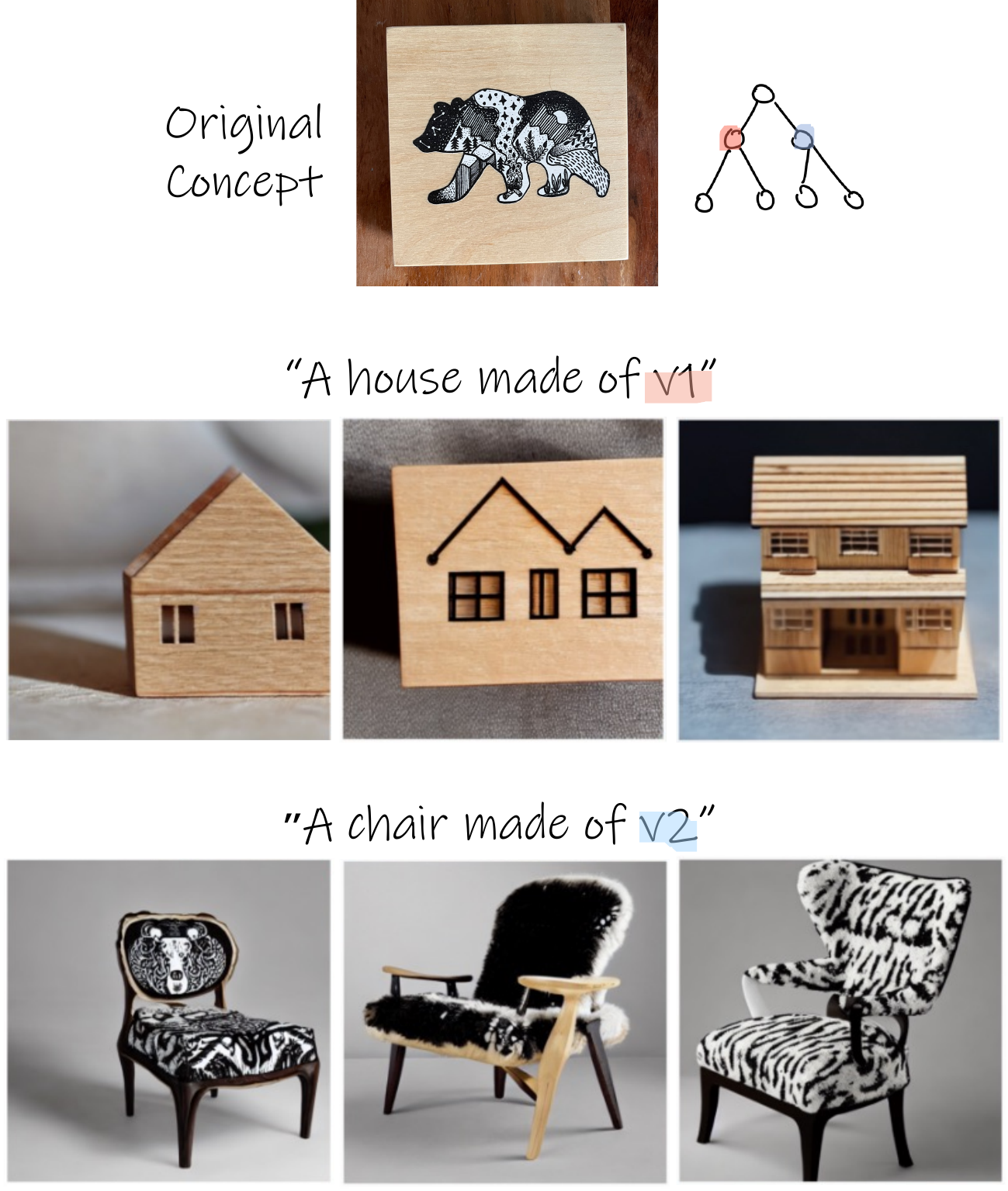}
    \caption{Combining the learned aspects in natural sentences to produce aspect-based variations.
    The original concept is shown on top, along with an illustration of the chosen aspects from the tree in \Cref{fig:teaser}.
    Below are three random images generated by a pretrained text-to-image model, conditioned on the prompts above.}
    \label{fig:text_based}
\end{figure}

We model the exploration space using a binary tree, where each node in the tree is a newly learned vector embedding in the textual latent space of a pretrained text-to-image model, representing different aspects of the original concept. 
A tree provides an intuitive structure to separate and navigate the different aspects of a given concept. Each level allows to find more aspects of the concepts in the previous level. In addition, each node by itself contains a plethora of samples and can be used for exploration.
For example, in \Cref{fig:teaser}, the original concept is first decomposed into its dominant semantic aspects: the wooden saucer in \ap{v1} and the bear drawing in \ap{v2}, next, the bear drawing is further separated into the general concept of a bear in \ap{v3} and its unique texture in \ap{v4}. 

Given a small set of images depicting the concept of interest as input, we build the tree gradually. For each node, we optimize two child nodes at a time to match the concept depicted in their parent.
We also utilize a CLIP-based \cite{clip} consistency measurement, to ensure that the concepts depicted in the nodes are coherent and distinct.
The different aspects are learned \textit{implicitly}, without any external constraint regarding the type of separation (such as shape or texture). 
As a result, unexpected concepts can emerge in the process and be used as inspiration for new design ideas.
For example the learned aspects can be integrated into existing concepts by combining them in natural language sentences passed to a pretrained text-to-image model (see \Cref{fig:text_based}).
They can also be used to create new concepts by combining different aspects of the same tree (intra-tree combination) or across different trees (inter-tree combination).

We provide many visual results applied to various challenging concepts. We demonstrate the ability of our approach to find different aspects of a given concept, explore and discover new concepts derived from the original one, thereby inspiring the generation of new design ideas.

\section{Previous Work}
\paragraph{Design and Modeling Inspiration}
Creativity has been studied in a wide range of fields \cite{Runco2012TheSD,Bonnardel2005TowardsSE,Amabile1996CreativityIC,Elhoseiny2019CreativityIZ,Kantosalo2014FromIT}, and although defining it exactly is difficult, some researchers have suggested that it can be described as the act of evoking and recombinating information from previous knowledge to generate new properties \cite{Bonnardel2005TowardsSE, WILKENFELD200121}.
It is essential, however, to be able to associate ideas in order to generate original ideas rather than just mimicking prior work \cite{Brown2008GUIDINGCD}. Experienced designers and artists are more adept at connecting disparate ideas than novice designers, who need assistance in the evocation process \cite{Bonnardel2005TowardsSE}. By reviewing many exemplars, designers are able to gain a deeper understanding of design spaces and solutions \cite{ECKERT2000523}. 
In the field of human-computer interaction, a number of studies have been conducted to develop tools and software to assist designers in the process of ideation \cite{ImageSense2020,MoodCubes2022,Koch2019MayAD,MetaMap2021}. 
They are focused on providing better tools for collecting, arranging, and searching visual and textual data, often collected from the web. In contrast, our work focuses on extracting different aspects of a given visual concept and \textit{generating} new images for inspiration.

Our work is close to a line of work utilizing evolutionary algorithms to inspire users' creativity \cite{CohenOr2015FromIM,Xu2012FitAD,Averkiou2014ShapeSynthPM}. However, they mostly work in the field of 3D content generation and do not decompose different aspects from existing concepts.

\paragraph{Large Language-Vision Models}
With the recent advancement of language-vision models \cite{clip} and diffusion models ~\cite{ramesh2022hierarchical, nichol2021glide, rombach2022highresolution}, the field of image generation and editing has undergone unprecedented evolution.
These models have been trained on millions of images and text pairs and have shown to be effective in performing challenging vision related tasks \cite{SegDiff,Avrahami_2022_CVPR,gal2021stylegannada,Patashnik_2021_ICCV,sheynin2022knndiffusion}.
Furthermore, the strong visual and semantic priors of these models have also been demonstrated to be effective for artistic and design tasks \cite{TianEvolution2021,vinker2022clipasso,clipascene,midjourney,Oppenlaender2022TheCO}.
In our work, we demonstrate how these models can be used to decompose and transform existing concepts into new ones in order to inspire the development of new ideas.

\paragraph{Personalization}
Personalized text-to-image generation has been introduced recently \cite{gal2022textual,ruiz2023dreambooth,kumari2022customdiffusion,hu2021lora}, with the goal of creating novel scenes based on user provided unique concepts.
In addition to demonstrating unprecedented quality results, these technologies enabled intuitive editing, made design more accessible, and attracted interest even beyond the research community.
We utilize these ideas to facilitate the ideation process of designers and common users, by learning different visual aspects of user-provided concepts.

Current personalization methods either optimize a set of embeddings to describe the concept \cite{gal2022textual}, or modify the denoising network to tie a rarely used word embedding to the new concept \cite{ruiz2023dreambooth}.
While the latter provides more accurate reconstruction and is more robust, it uses much more memory and requires a model for each object.
In this regard, we choose to rely on the approach presented in \cite{gal2022textual}.
It is important to note that our goal is to capture multiple \textit{aspects} of the given concept, and not to improve the accuracy of reconstruction as in \cite{Wei2023ELITEEV,Tewel2023KeyLockedRO,Gal2023DesigningAE,Voynov2023PET,Shi2023InstantBoothPT,Han2023SVDiffCP}.

\section{Preliminaries}
\paragraph{Latent Diffusion Models.}
Diffusion models are generative models trained to learn data distribution by gradually denoising a variable sampled from a Gaussian distribution.

In our work, we use the publicly available text-to-image Stable Diffusion model \cite{rombach2022highresolution}. 
Stable Diffusion is a type of a latent diffusion model (LDM), where the diffusion process is applied on the latent space of a pretrained image autoencoder. The encoder $\mathcal{E}$ maps an input image $x$ into a latent vector $z$, and the decoder $\mathcal{D}$ is trained to decode $z$ such that $\mathcal{D}(z)\approx x$.
As a second stage, a denoising diffusion probabilistic model (DDPM) \cite{ddpm} is trained to generate codes within the learned latent space.
At each step during training, a scalar $t\in\{1,2,...T\}$ is uniformly sampled and used to define a noised latent code $z_t = \alpha_t z + \sigma_t \epsilon$, where $\epsilon \sim \mathcal{N}(0, I)$ and $\alpha_t, \sigma_t$ are terms that control the noise schedule, and are functions of the diffusion process time $t$.
The denoising network $\epsilon_{\theta}$ which is based on a UNet architecture \cite{unet2015}, receives as input the noised code $z_t$, the timestep $t$, and an optional condition vector $c(y)$, and is tasked with predicting the added noise $\epsilon$.
The LDM loss is defined by:
\begin{equation}
    \mathcal{L}_{LDM} = \mathbb{E}_{z\sim\mathcal{E}(x),y,\epsilon\sim\mathcal{N}(0,1),t} \left [ || \epsilon - \epsilon_\theta(z_t, t, c(y)) ||_2^2 \right ]
    \label{eq:ldm_loss}
\end{equation}
For text-to-image generation the condition $y$ is a text input and $c(y)$ represents the text embedding.
At inference time, a random latent code $z_T \sim \mathcal{N}(0, I)$ is sampled, and iteratively denoised by the trained $\epsilon_{\theta}$ until producing a clean $z_0$ latent code, which is passed through the decoder $D$ to produce the image $x$. 

We next discuss the text encoder and the inversion space.

\begin{figure*}
    \centering
    \includegraphics[width=0.95\textwidth]{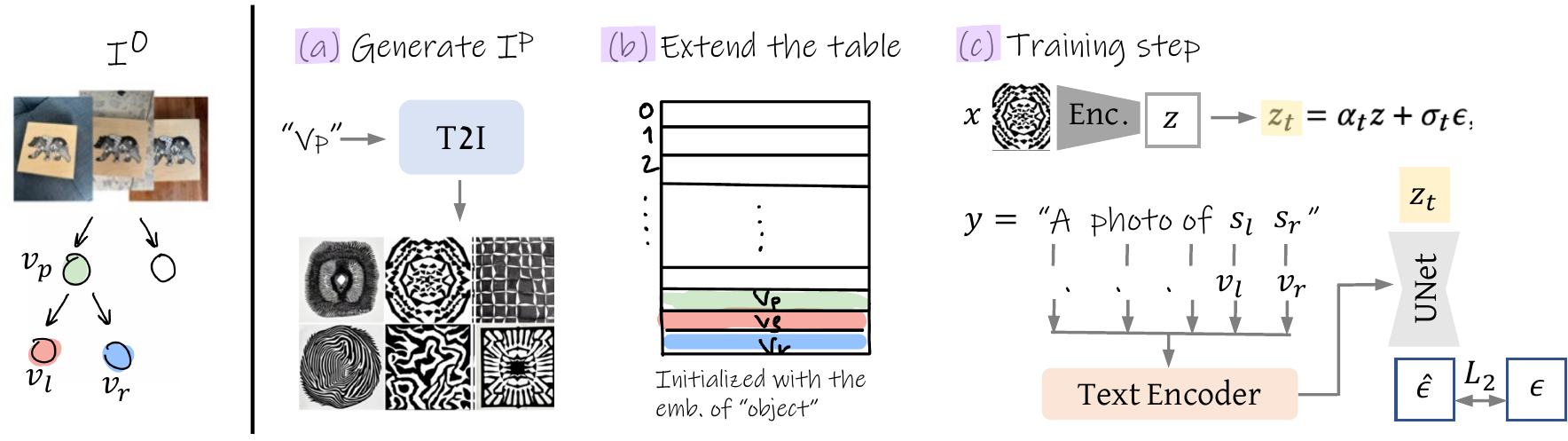}
    \vspace{-0.3cm}
    \caption{High level pipeline of the \ap{binary reconstruction} stage. We optimize two sibling nodes $v_l, v_r$ at a time (marked in red and blue). (a) We first generate a small training set of images $I^p$ depicting the concept in the parent node using a pretrained text-to-image model (T2I). At the root, we use the original set of images $I^0$. (b) We then extend the existing dictionary by adding the two new vectors, initialized with the embedding of the word \ap{object}. (c) Lastly, we optimize $v_l, v_r$ w.r.t. the LDM loss (see details in the text).}
    \vspace{-0.3cm}
    \label{fig:pipeline1}
\end{figure*}

\paragraph{Text embedding.}
Given a text prompt $y$, for example ``A photo of a cat'', the sentence is first converted into tokens, which are indexed into a pre-defined dictionary of vector embeddings.
The dictionary is a lookup table that connects each token to a unique embedding vector.
After retrieving the vectors for a given sentence from the table, they are passed to a text transformer, which processes the connections between the individual words in the sentence and outputs $c(y)$. 
The output encoding $c(y)$ is then used as a condition to the UNet in the denoising process.
We denote words with $S$, and the vector embeddings from the lookup table with $V$.

\paragraph{Textual Inversion}
We rely on the general framework proposed by \cite{gal2022textual}, who choose the embedding space of $V$ as the target for inversion. They formulate the task of inversion as fitting a new word $s^*$ to represent a personal concept, depicted by a small set of input images provided by the user.
They extend the predefined lookup table with a new embedding vector $v_*$ that is linked to $s^*$. 
The vector $v_*$ is often initialized with the embedding of an existing word from the dictionary that has some relation to the given concept, and then optimized to represent the desired personal concept. 
This process can be thought of as ``injecting'' the new concept into the vocabulary.
The vector $v_*$ is optimized w.r.t. the LDM loss in \Cref{eq:ldm_loss} over images sampled from the input set.
At each step of optimization, a random image $x$ is sampled from the set, along with a neutral context text $y$, derived from the CLIP ImageNet templates~\cite{radford2021learning} (such as \ap{A photo of $s^*$}).
Then, the image $x$ is encoded to $z=\mathcal{E}(x)$ and noised w.r.t.\ a randomly sampled timestep $t$ and noise $\epsilon$: $z_t = \alpha_t z + \sigma_t\epsilon$.
The noisy latent image $z_t$, timestep $t$, and text embedding $c(y)$ are then fed into a pretrained UNet model which is trained to predict the noise $\epsilon$ applied w.r.t.\ the conditioned text and timestep.
This way, $v_*$ is optimized to describe the object depicted in the small training set of images.
\section{Method}
Given a small set of images $I^0 = \{I_1^0 ... I_m^0\}$ depicting the desired visual concept, our goal is to construct a rich visual exploration space expressing different aspects of the input concept.

We model the exploration space as a binary tree, whose nodes $V = \{v_1 .. v_n\}$ are learned vector embeddings corresponding to newly discovered words $S = \{s_1 .. s_n\}$ added to the predefined dictionary, representing different aspects of the original concept. 
These newly learned words are used as input to a pretrained text-to-image model \cite{rombach2022highresolution} to generate a rich variety of image examples in each node. 
We find a binary tree to be a suitable choice for our objective, because of the ease of visualization, navigation, and the quality of the sub-concepts depicted in the nodes (see supplemental file for further analysis).

\subsection{Tree Construction}
The exploration tree is built gradually as a binary tree from top to bottom, where we iteratively add two new nodes at a time.
To create two child nodes, we optimize new embedding vectors according to the input image-set generated from the concept depicted in the parent node.
During construction, we define two requirements to encourage the learned embeddings to follow the tree structure: (1) \textbf{Binary Reconstruction} each pair of children nodes together should encapsulate the concept depicted by their parent node, and (2) \textbf{Coherency} each individual node should depict a coherent concept which is distinct from its sibling. Next, we describe the loss functions and procedures designed to follow these requirements.

\begin{figure}[t]
    \centering
    \includegraphics[width=0.92\linewidth]{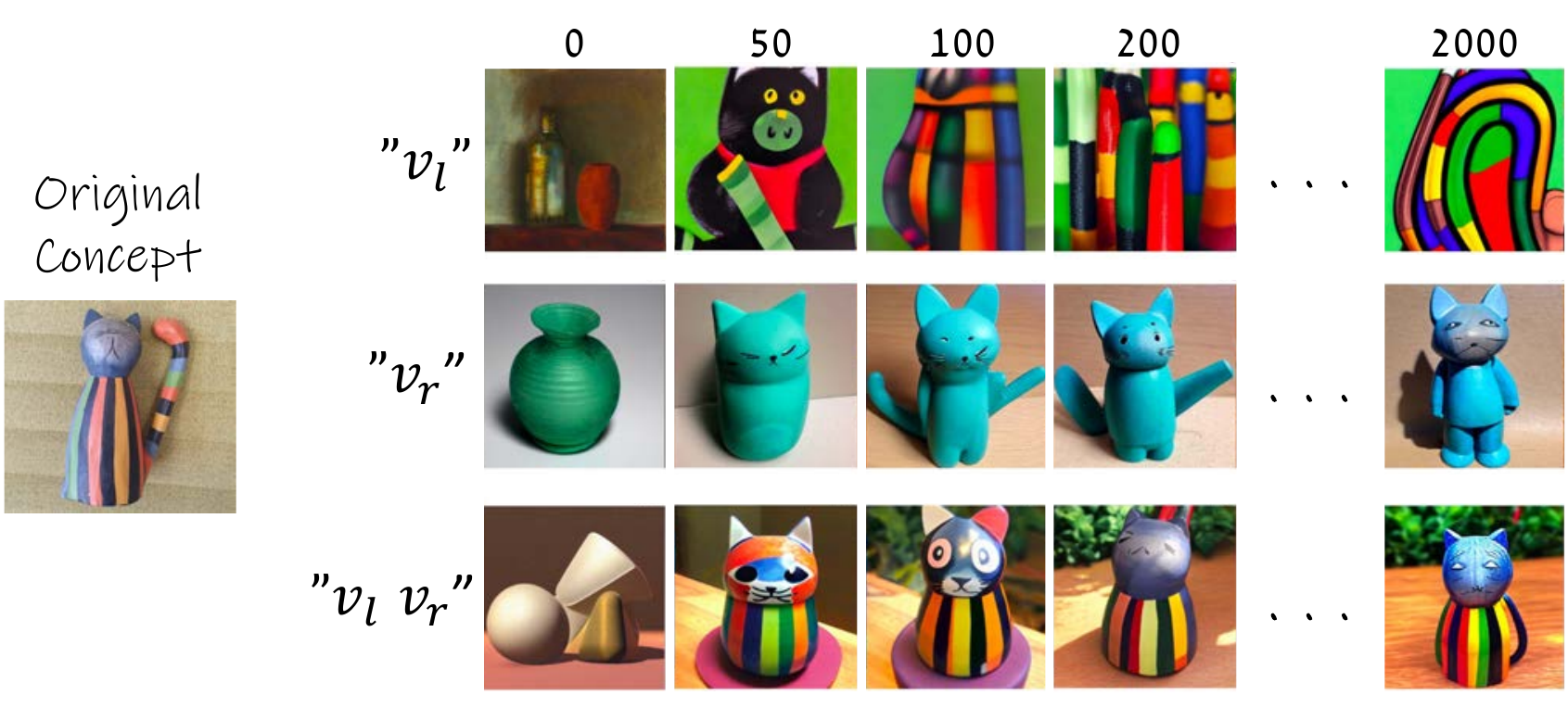}
    \caption{Optimization iterations. The embedding of both children nodes $v_l, v_r$ are initilized with the word ``object''. During iterations, they gradually depict two aspects of the original concept. Note that using both embedding together reconstructs the original parent concept.}
    \label{fig:iter_example}
\end{figure}

\vspace{-0.1cm}
\paragraph{\textbf{Binary Reconstruction}}
We use the reconstruction loss suggested in \cite{gal2022textual}, with some modifications tailored to our goal.
The procedure is illustrated in \Cref{fig:pipeline1} -- in each optimization phase, our goal is to learn two vector embeddings $v_l, v_r$ corresponding to the left and right sibling nodes, whose parent node is marked with $v_p$ (illustrated in \Cref{fig:pipeline1}, left).
We begin with generating a new small training set of images $I^p = \{I_1^p ... I_{10}^p\}$, reflecting the concept depicted by the vector $v_p$ (\Cref{fig:pipeline1}a). At the root, we use the original set of images $I^0$.
Next, we extend the current dictionary by adding two new vector embeddings $v_l, v_r$, corresponding to the right and left children of their parent node $v_p$ (\Cref{fig:pipeline1}b). 
To represent general concepts, the newly added vectors are initialized with the embedding of the word \ap{object}.
At each iteration of optimization (\Cref{fig:pipeline1}c), an image $x$ is sampled from the set $I^p$ and encoded to form the latent image $z = \mathcal{E}(x)$.
A timestep $t$ and a noise $\epsilon$ are also sampled to define the noised latent $z_t = \alpha_t z + \sigma_t \epsilon$ (marked in yellow).
Additionally, a neutral context text $y$ is sampled, containing the new placeholder words in the following form ``A photograph of $s_l$ $s_r$''.
The noised latent $z_t$ is fed to a pretrained Stable Diffusion UNet model $\epsilon_\theta$, conditioned on the CLIP embedding $c(y)$ of the sampled text, to predict the noise $\epsilon$.
The prediction loss is backpropagated w.r.t. the vector embeddings $v_l, v_r$:
\begin{equation}
   \{v_l, v_r\} = \argmin_v \mathbb{E}_{z\sim\mathcal{E}(x), y, \epsilon \sim \mathcal{N}(0, 1), t }\Big[ \Vert \epsilon - \epsilon_\theta(z_{t},t, c(y)) \Vert_{2}^{2}\Big] .
    \label{eq:v_opt}
\end{equation}
This procedure encourages $v_l, v_r$ together to express the visual concept of their parent depicted in the set $I^p$.
\Cref{fig:iter_example} illustrates how the two embeddings begin by representing the word \ap{object}, and gradually converge to depict two aspects of the input concept. 

We use the timestep sampling approach proposed in ReVersion \cite{huang2023reversion}, which skews the sampling distribution so that a larger $t$ is assigned a higher probability, according to the following importance sampling function:
\begin{equation}
   f(t) = \frac{1}{T}(1 - \alpha \cos \frac{\pi t}{T}).
\end{equation}
We set $\alpha = 0.5$. We find that this sampling approach improves stability and content separation. This choice is further discussed in the supplementary file.

\begin{figure}
    \centering
    \includegraphics[width=0.94\linewidth]{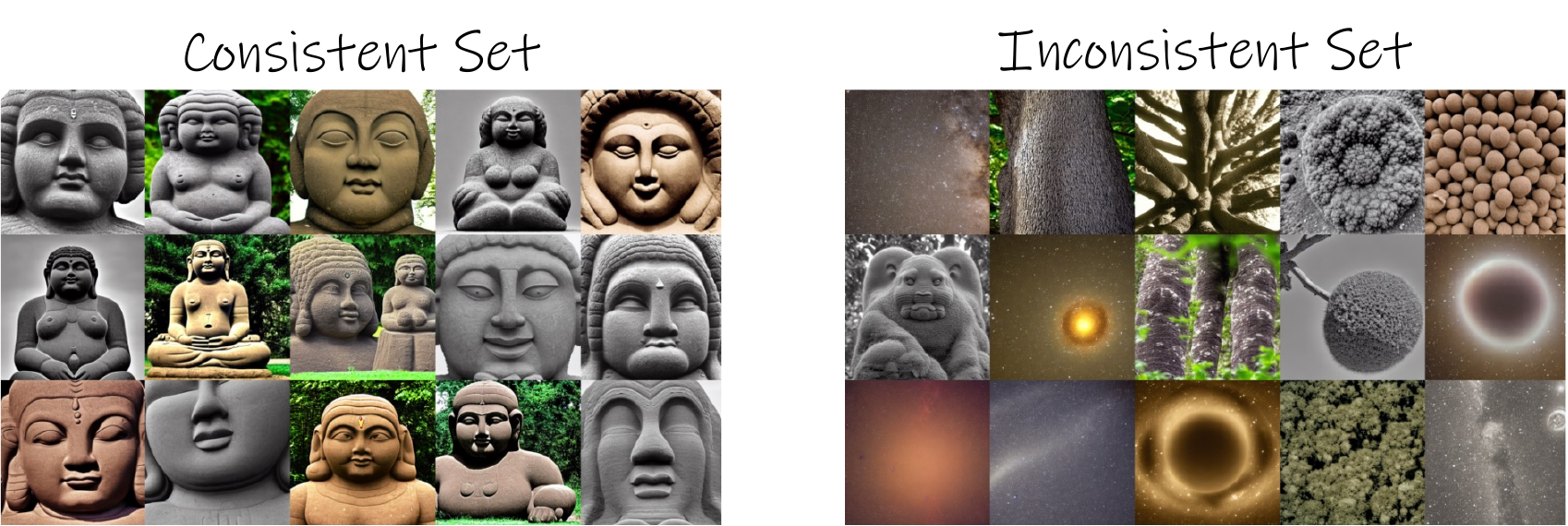}
    \caption{We demonstrate two sets of random images generated from two different vector embeddings. An example of a consistent set can be seen on the left, where the concept depicted in the node is clear. We show an inconsistent set on the right, where images appear to depict multiple concepts.}
    \label{fig:consistency_buddha}
\end{figure}

\paragraph{\textbf{Coherency}}
The resulting pair of embeddings described above together often capture the parent concept depicted in the original images well.
However, the images produced by each embedding individually may not always reflect a logical sub-concept that is coherent to the observer.

We find that such incoherent embeddings are frequently characterized by inconsistent appearance of the images generated from them, i.e., it can be difficult to identify a common concept behind them. For example, in \Cref{fig:consistency_buddha} the concept depicted in the set on the right is not clear, compared to the set of images on the left. 

This issue may be related to the observation that textual inversion often results in vector embedding outside of the distribution of common words in the dictionary, affecting editability as well \cite{Voynov2023PET}.
It is thus possible that embeddings that are highly unusual may not behave as ``real words'', thereby producing incoherent visual concepts.
In addition, textual-inversion based methods are sometimes unstable and depend on the seed and iteration selection.

To overcome this issue we define a consistency test, which allows us to filter out incoherent embeddings. We begin by running the procedure described above to find $v_l, v_r$ using $k$ different seeds in parallel for a sufficient number of steps (in our experiments we found that k=4 and 200 steps are sufficient since at that point the embeddings have already progressed far enough from their initialization word ``object'' as seen in \Cref{fig:iter_example}).

This gives us an initial set of $k$ pairs of vector embeddings $V_{s} = \{v_l^i, v_r^i\}_{i=1}^{k}$.
For each vector $v \in V_s$ we generate a random set $I^v$ of 40 images using our pre-trained text-to-image model.
We then use a pretrained CLIP Image encoder \cite{clip}, to produce the embedding $CLIP(I_i^v)$ of each image in the set.

We define the consistency of two sets of images $I^a, I^b$ as follows:

\begin{equation} \label{eq:const}
    \begin{split}
    \mathcal{C}(I^{a}, I^{b}) & = mean_{I_i^{a}\in{I^{a}} ,I_j^{b}\in{I^{b}, I_i^{a} \neq I_j^{b}} } \\
    & (sim(CLIP(I_i^{a}), CLIP(I_j^{b}))).
    \end{split}
\end{equation}

\begin{figure}[t]
    \centering
    \includegraphics[width=0.9\linewidth]{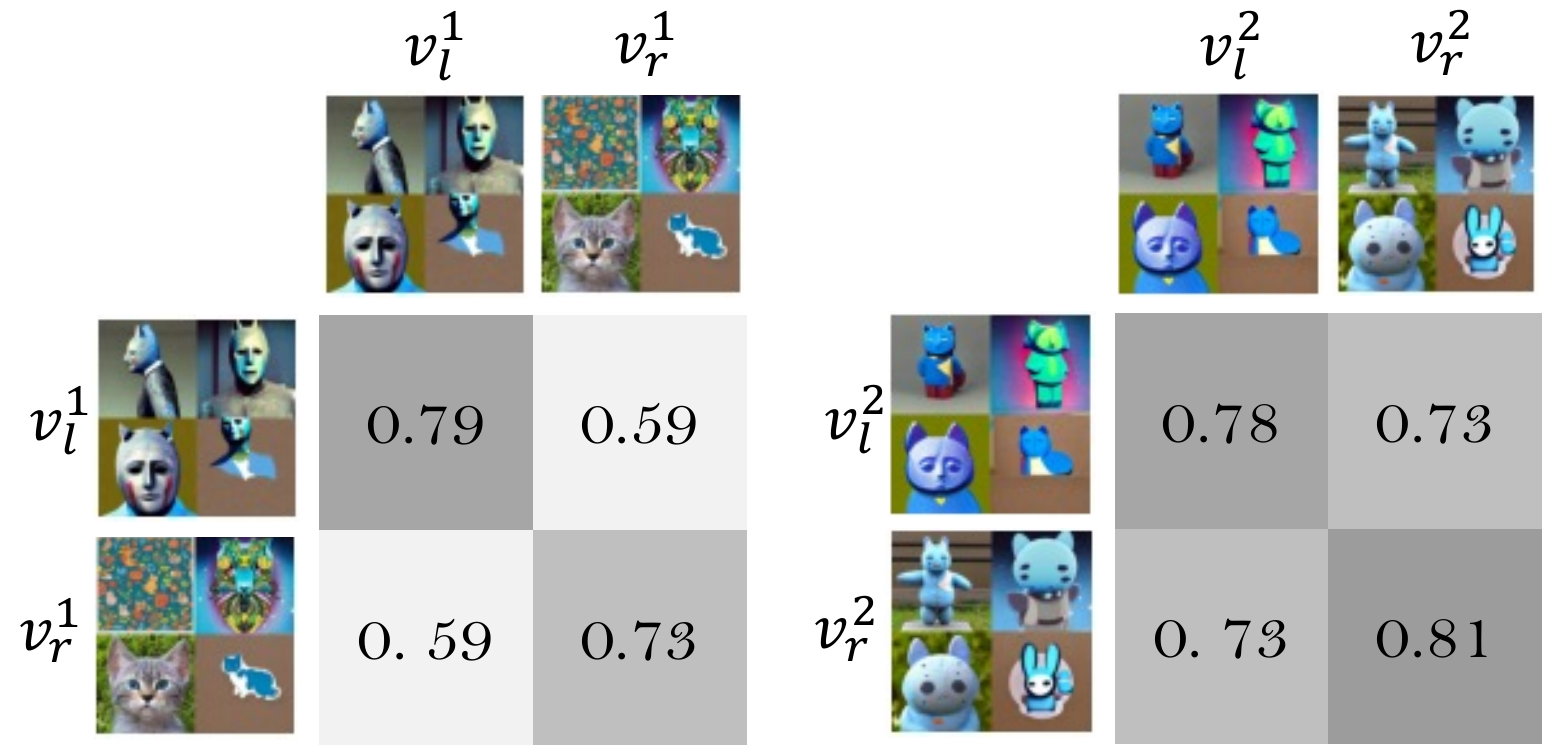}
    \vspace{-0.2cm}
    \caption{Consistency scores matrix between image sample sets of nodes. The seed selection process favors pairs of siblings that have a high consistency score within themselves, and low consistency score between each other. In this example, the left pair is better than the right.}
    \vspace{-0.1cm}
    \label{fig:coherency_matrix}
\end{figure}

Note that $|\mathcal{C}(I^{a}, I^{b})| \leq 1$ because $sim(x,y) = \frac{x\cdot y}{||x||\cdot||y||}$ is the cosine similarity between a pair of CLIP embedding of two different images. This formulation is motivated by the observation that if a set of images depicts a certain semantic concept, their vector embedding in CLIP's latent space should be relatively close to each other. 
Ideally, we are looking for pairs in which each node is coherent by itself, and in addition, two sibling nodes are distinct from each other.
We therefore choose the pair of tokens $\{v_l^*, v_r^*\}\in V_{s}$ as follows:

\begin{equation} \label{eq:consistency}
    \begin{split}
    \{v_l^*, v_r^*\} & = \argmax_{\{v_l^i, v_r^i\} \in V_s} \big[ C_l^i + C_r^i + \\
     & = (min(C_l^i, C_r^i) - \mathcal{C}(I^{v_l^i}, I^{v_r^i})) \big],
    \end{split}
\end{equation}

where $ C_l^i = \mathcal{C} (I^{v_l^i}, I^{v_l^i}), C_r^i = \mathcal{C}(I^{v_r^i}, I^{v_r^i})$.
Note that we do not consider the absolute cross consistency score $\mathcal{C}(I^{v_l^i}, I^{v_r^i})$, but we compute its relative difference from the node with the minimum consistency.
We demonstrate this procedure in \Cref{fig:coherency_matrix}.
We optimized two pairs of sibling nodes $\{v_l^1, v_r^1\}, \{v_l^2, v_r^2\}$ using two seeds, w.r.t. the same parent node. Each matrix illustrates the consistency scores $C_l^i, \mathcal{C}(I^{v_l^i}, I^{v_r^i}), C_r^i$ obtained for the sets of images of each seed.
In both cases, the scores on the diagonal are high, which indicates that each set is consistent within itself.
While the sets on the right obtained a higher consistency score within each node, they also obtained a relatively high score across the nodes (0.73), which means they are not distinct enough.

After selecting the optimal seed, we continue the optimization of the chosen vector pair w.r.t. the reconstruction loss in \Cref{eq:v_opt} for $1500$ iterations.

\begin{figure*}
    \centering
    \includegraphics[width=0.97\linewidth]{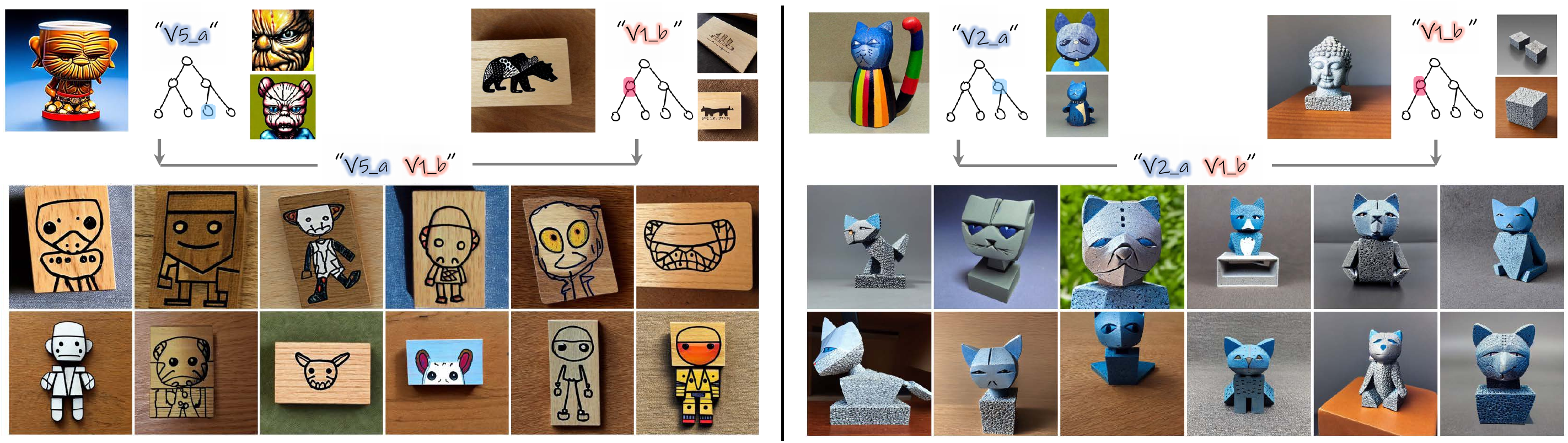}
    \caption{Examples of inter-tree combinations. We use our method to produce trees for the four concepts depicted in the first row. We then combine aspects from different trees to generate a set of inter-tree combinations (the chosen aspects are shown next to each concept).}
    \label{fig:mug_bear_comb}
\end{figure*}

\section{Results}
\begin{figure*}
    \centering
    \includegraphics[width=0.9\textwidth]{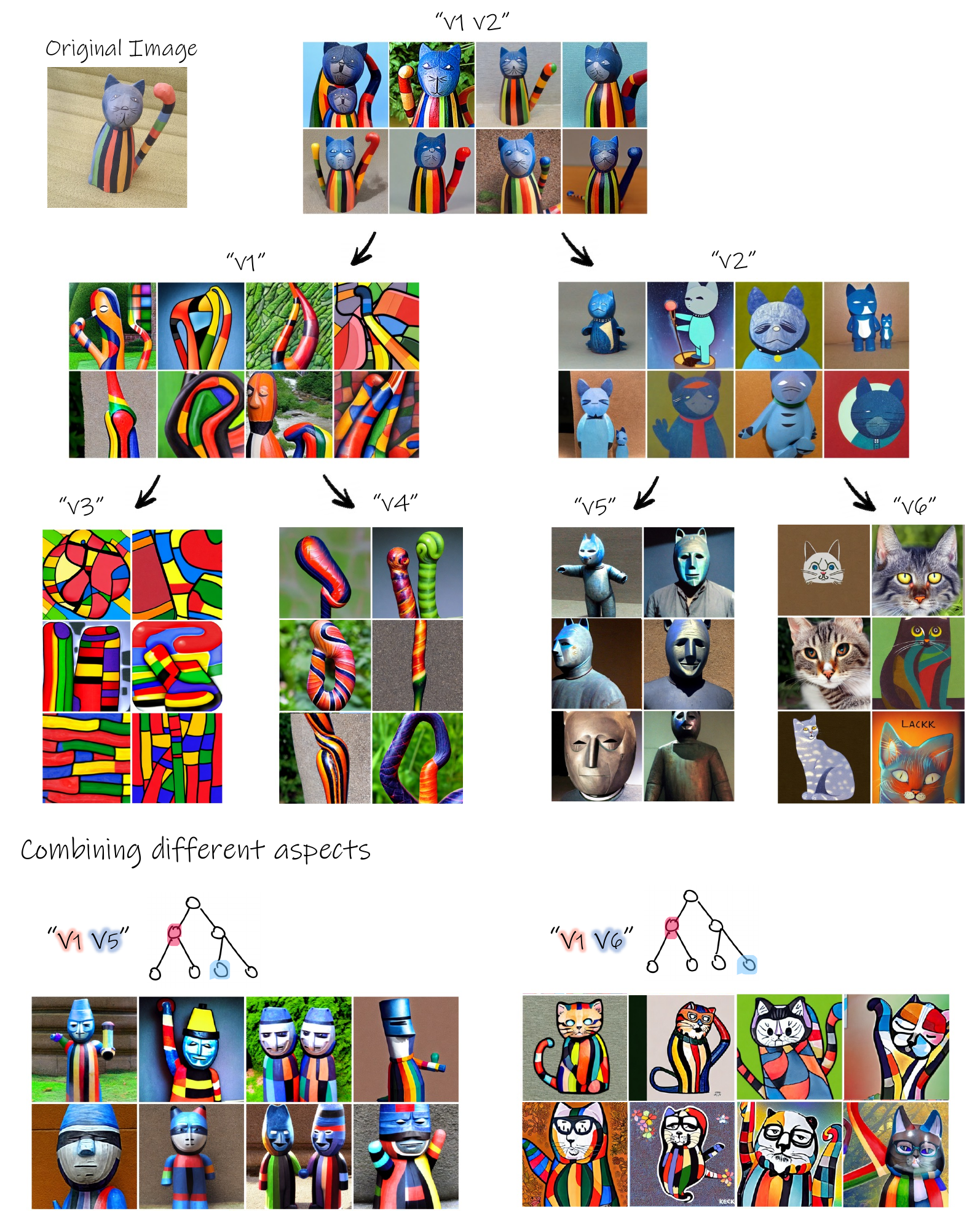}
    \caption{Exploration tree for the cat sculpture. At the bottom we show examples of possible intra-tree combinations.}
    \label{fig:cat_tree}
\end{figure*}

\begin{figure*}
    \centering
    \includegraphics[width=0.9\textwidth]{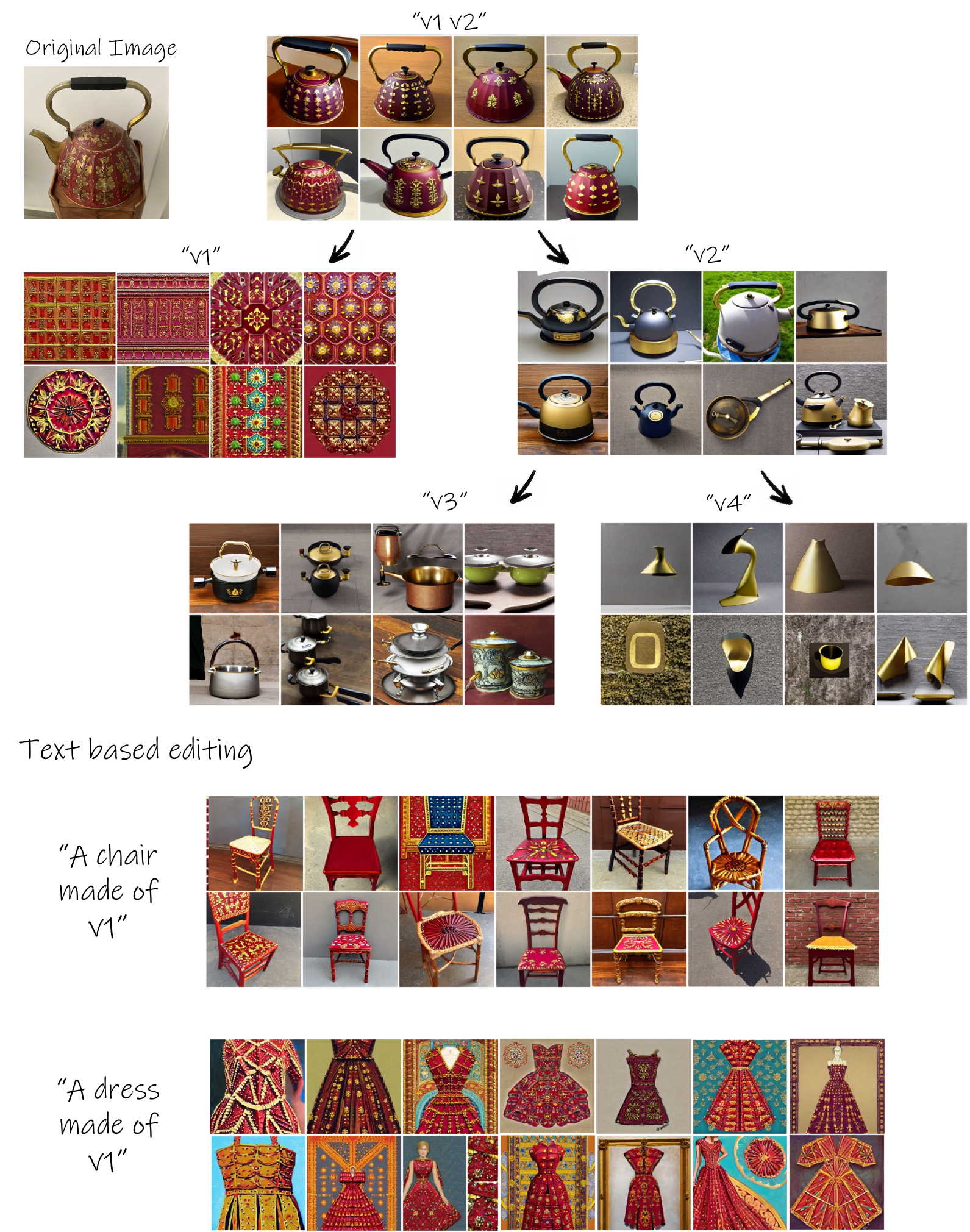}
    \caption{Exploration tree for a decorated teapot. At the bottom we show examples of possible text-based generation.}
    \label{fig:red_teapot}
\end{figure*}

In \Cref{fig:teaser,fig:cat_tree,fig:red_teapot}, we show examples of possible trees. 
For each node in the tree, we use its corresponding placeholder word as an input to a pretrained text-to-image model \cite{rombach2022highresolution}, to generate a set of random images.
These images have been generated without any prompt engineering or additional words within the sentence, except for the word itself. For clarity, we use the notion $"v"$ next to each set of images, illustrating that the presented set depicts the concept learned in that node.
As can be seen, the learned embeddings in each node capture different elements of the original concept, such as the concept of a cat and a sculpture, as well as the unique texture in \Cref{fig:cat_tree}. 
The sub-concepts captured in the nodes follow the tree's structure, where the concepts are decomposed gradually, with two sibling nodes decomposing their parent node.
This decomposition is done \textit{implicitly}, without external guidance regarding the split theme.
For many more trees please see our supplementary file. 

\vspace{-0.11cm}
\subsection{Applications}
The constructed tree provides a rich visual exploration space for concepts related to the object of interest.
In this section we demonstrate how this space can be used for novel combination and exploration.

\noindent \textbf{Intra-tree combination} --
the generated tree is represented via the set of optimized vectors $V = \{v_1 .. v_n\}$. Once this set is learned we can use it to perform further exploration and conceptual editing \textit{within} the object's ``inner world''.
We can explore combinations of different aspects by composing sentences containing different subsets of $V$.
For example, in the bottom left area of \Cref{fig:cat_tree}, we have combined $v_1$ and $v_5$, which resulted in a variation of the original sculpture without the sub-concept relating to the cat (depicted in $v_6$). At the bottom right, we have excluded the sub-concept depicted in $v_5$ (related to a blue sculpture), which resulted in a new representation of a flat cat with the body and texture of the original object.

Such combinations can provide new perspectives on the original concept and inspiration that highlights only specific aspects.

\noindent \textbf{Inter-tree combination} --
it is also possible to combine concepts learned across different trees, since we only inject new words into the existing dictionary, and do not fine-tune the model's weights as in other personalization approaches \cite{ruiz2023dreambooth}.

To achieve this, we first build the trees independently for each concept and then visualize the sub-concepts depicted in the nodes to select interesting combinations.
In \Cref{fig:mug_bear_comb} the generated original concepts are shown on top, along with an illustration of the concepts depicted in the relevant nodes.
To combine the concepts across the trees, we simply place the two placeholder words together in a sentence and feed it into the pretrained text-to-image model.
As can be seen, on the left the concept of a \ap{saucer with a drawing} and the \ap{creature} from the mug are combined to create many creative and surprising combinations of the two.
On the right, the blue sculpture of a cat is combined with the stone depicted at the bottom of the Buddha, which together create new sculptures in which the Buddha is replaced with the cat.

\begin{figure}
    \centering
    \includegraphics[width=1\linewidth]{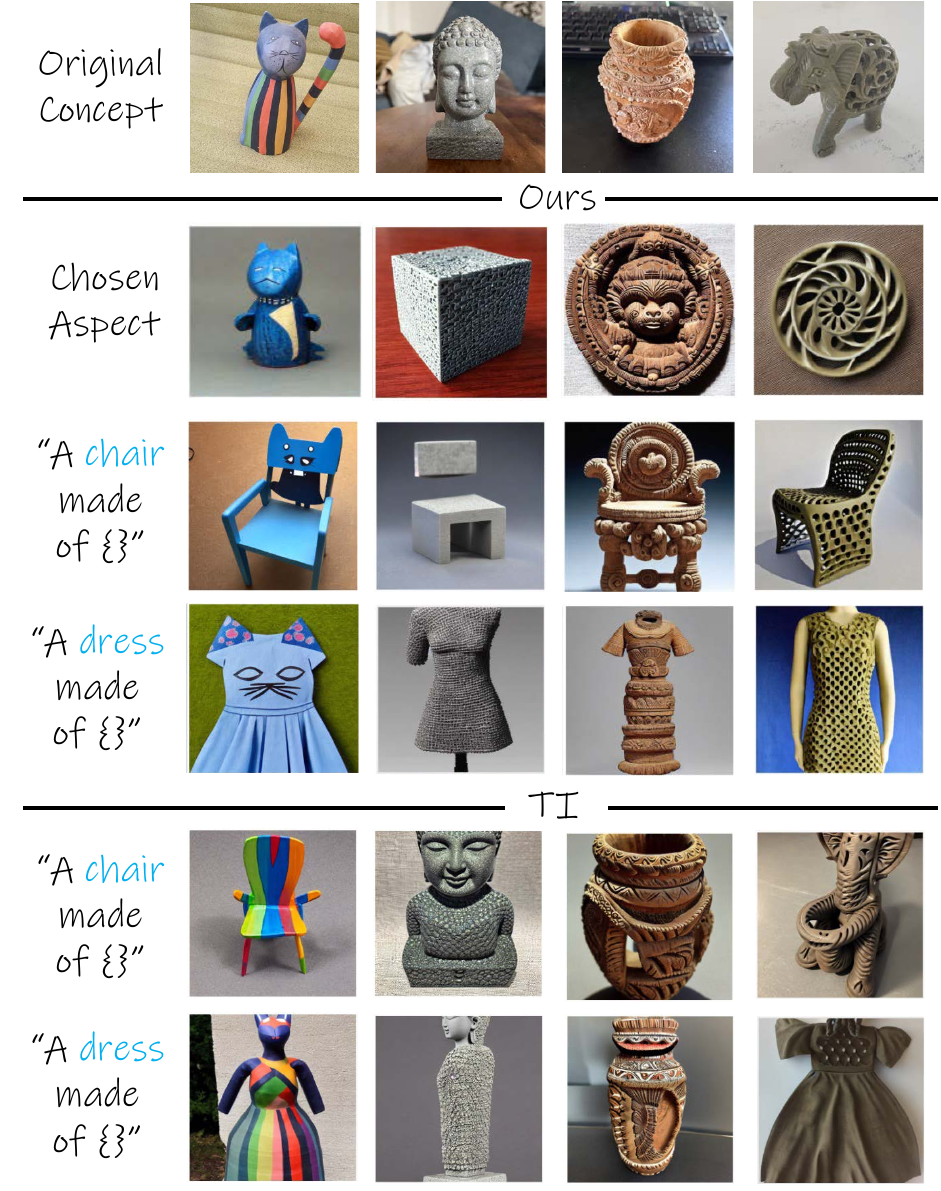}
    \vspace{-0.5cm}
    \caption{Combining the learned aspects in natural sentences to produce aspect-based variations.
    The original concepts are shown at the top. In the third and fourth rows are our text-based generation results applied with the aspects depicted in the second row. Under \ap{TI} we show image generation for the concepts in the first row (without our aspect decomposition approach), produced using \cite{gal2022textual}.}
    \vspace{-0.5cm}
    \label{fig:text_based_multiple}
\end{figure}

\noindent \textbf{Text-based generation} --
the placeholder words of the learned embeddings can be composed into natural language sentences to generate various scenes based on the learned aspects.
We illustrate this at the top of \Cref{fig:text_based_multiple}, where we integrate the learned aspects of the original concepts in new designs (in this case of a chair and a dress).
At the bottom of \Cref{fig:text_based_multiple}, we show the effect of using the learned vectors of the original concepts instead of specific aspects. We apply Textual Inversion (TI) \cite{gal2022textual} with the default hyperparameters to fit a new word depicting each concept, and choose a representative result.
The results suggest that without aspect decomposition, generation can be quite limited.
For instance, in the first column, both the dress and the chair are dominated by the texture of the sculpture, whereas the concept of a blue cat is almost ignored. Furthermore, TI may exclude the main object of the sentence (second and third columns), or the results may capture all aspects of the object (fourth column), thereby narrowing the exploration space.

\subsection{Evaluations}
\paragraph{Consistency Score Validation.}
We first show that our consistency test proposed in \Cref{eq:const} aligns well with human perception of consistency.
We conducted a perceptual study with $35$ participants in which we presented $15$ pairs of random image sets depicting sub-concepts of $9$ objects.
We asked participants to determine which of the sets is more consistent within itself in terms of the concept it depicts (an example of such a pair can be seen in \Cref{fig:consistency_buddha}).
We also measured the consistency scores for these sets using our CLIP-based approach, and compared the results. The CLIP-based scores matched the human choices in $82.3\%$ of the cases.

\vspace{-0.2cm}
\paragraph{Reconstruction and Separation.} We quantitatively evaluate our method's ability to follow the tree requirements of reconstruction and sub-concept separation. 
We collected a set of $13$ concepts ($9$ from existing personalization datasets \cite{gal2022textual,kumari2022customdiffusion}, and $4$ new concepts from our dataset), and generated $13$ corresponding trees.
Note that we chose concepts that are complex enough and have the potential to be divided into different aspects (we discuss this in the limitations section).
For each pair of sibling nodes $v_l, v_r$ and their parent node $v_p$, we produced their corresponding sets of images -- $I^{v_l}, I^{v_r}, I^{v_p}$ (where for nodes in the first level we used the original set of images $I^0$ as $I^{v_p}$). We additionally produced the set $I^{v_l v_r}$, depicting the joint concept learned by two sibling nodes.

We first compute $\mathcal{C}(I^{v_p}, I^{v_l v_r})$ to measure the quality of reconstruction, i.e., that two sibling nodes together represent the concept depicted in their parent node. The average score obtained for this measurement is $0.8$, which suggests that on average, the concept depicted by the children nodes together is consistent with that of their parent node.
Second, we measure if two sibling nodes depict distinct concepts by using $\mathcal{C}(I^{v_l}, I^{v_r})$. The average score obtained was $0.59$, indicating there is larger separation between siblings, but they are still close.

\paragraph{Aspects Relevancy.}
We assess the ability of our method to encode different aspects connected to the input concept via a perceptual study.
We chose 5 objects from the dataset above, and 3 random aspects for each object.
We presented participants with a random set of images depicting one aspect of one object at a time. We asked the participants to choose the object they believe this aspect originated from, along with the option `none'. In total we collected answers from $35$ participants, and achieved recognition rates of $87.8\%$. 
These evaluations demonstrate that our method can indeed separate a concept into \emph{relevant} aspects, where each new sub-concept is \emph{coherent}, and the binary tree structure is valid - i.e., the combination of two children can \emph{reconstruct} the parent concept. 

\begin{figure}
    \centering
    \includegraphics[width=1\linewidth]{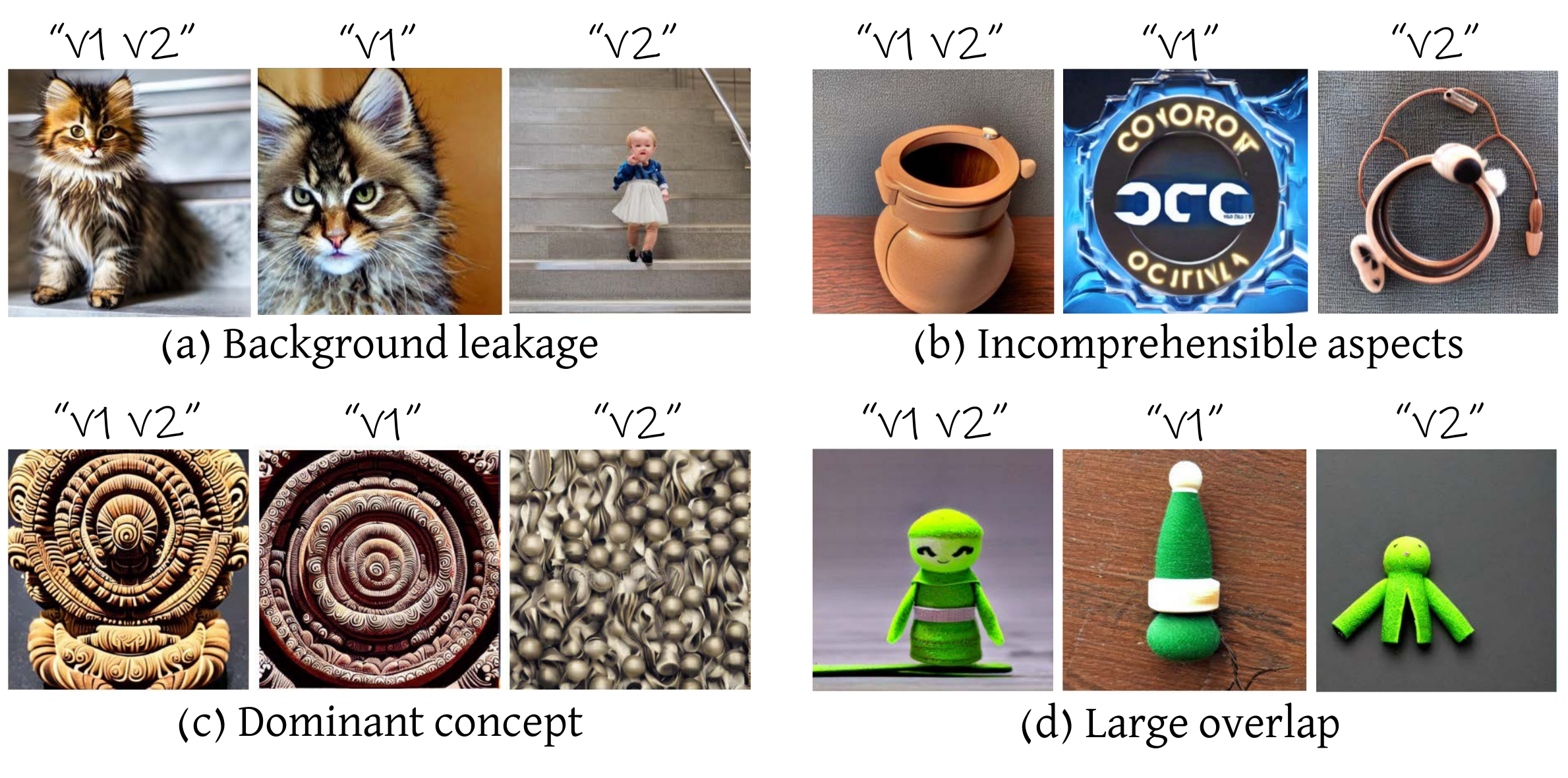}
    \vspace{-0.5cm}
    \caption{We demonstrate four general cases of decomposition failure.}
    \vspace{-0.5cm}
    \label{fig:limitations}
\end{figure}

\section{Limitations}
Our method may fail to decompose an input concept. We divide the failure cases into four general categories illustrated in \Cref{fig:limitations}:

\noindent (1) Background leakage - the training images should be taken from different perspectives and with varying backgrounds (this requirement also exists in \cite{gal2022textual}). When images do not meet these criteria, one of the sibling nodes often captures information from the background instead of the object itself. 

\noindent (2) Incomprehensible aspects - some separations may not satisfy clear, interesting, aesthetic, or inspiring aspects, even when the coherency principle holds.

\noindent (3) Dominant sub-concept - we illustrate this in \Cref{fig:limitations}c, where we show a split on the second level of the concept depicted under \ap{$v_1 v_2$}. As shown, v1 has dominated the information, so even if the coherency term is held, decomposition to two sub-concepts has not really been achieved.

\noindent (4) Large overlap when two aspects share information -- we illustrate this in \Cref{fig:limitations}d, which is a split of the second level, where both concepts depicted in v1 and v2 appear to share too similar.

We hope that such limitations could be resolved in the future using additional regularization terms in the optimization process or through the development of more robust personalization methods.

Additionally, our method can have difficulties to create deeper trees and nodes with more than two children (see examples in supplemental file). 
Currently, we stop the process when sub-concepts become too simple or incoherent. 
This could be the result of the new embeddings drifting towards out-of-distribution codes. Further investigation is needed in this subject.
Currently the time for decomposing a node can reach up to approximately $40$ minutes on a single A100 GPU. However, as textual inversion optimization techniques will progress, so will our method.

\section{Conclusions}
We presented a method to implicitly decompose a given visual concept into various aspects to construct an inspiring visual exploration space.
Our method can be used to generate numerous representations and variations of a certain subject, to combine aspects across objects, as well as to use these aspects as part of natural language sentences that drive visual generation of novel concepts.

The aspects are learned implicitly, without external guidance regarding the type of separation.
This implicit approach also provides another small step in revealing the rich latent space of large vision-language models, allowing surprising and creative representations to be produced.
We demonstrated the effectiveness of our method on a variety of challenging concepts. We hope our work will open the door to further research aimed at developing and improving existing tools to assist and inspire designers and artists.

\paragraph{Acknowledgements}
We thank Rinon Gal, Kfir Aberman, and Yael Pritch for their early feedback and insightful discussions.

\null
\newpage
\null
\newpage
\null
\newpage
\null
\newpage

{\small
\bibliographystyle{ieee_fullname}
\bibliography{bibliography}
}

\clearpage
\appendix
\appendixpage
\addcontentsline{toc}{section}{} %
\part{} %
\vspace{-3em}
\parttoc %
\section{Implementation details}
We rely on the diffusers \cite{von-platen-etal-2022-diffusers} implementation of Textual Inversion \cite{gal2022textual}, based on Stable Diffusion v1.5 text-to-image model~\cite{radford2021learning}.
We used the default training parameters provided in this implementation, except for changing the batch size to $2$ (which scales the learning rate to $0.004$ respectively).
We used four different seeds $\{0,1000,1234,111\}$ for each sibling nodes optimization.
To generate the set of $10$ images for each new node we first generated a random set of 40 images, and used our proposed CLIP consistency measurement to choose a subset of 10 images that are most consistent with each other.
Our code will be made available to facilitate future research.
\begin{figure*}
    \centering
    \includegraphics[width=0.85\linewidth]{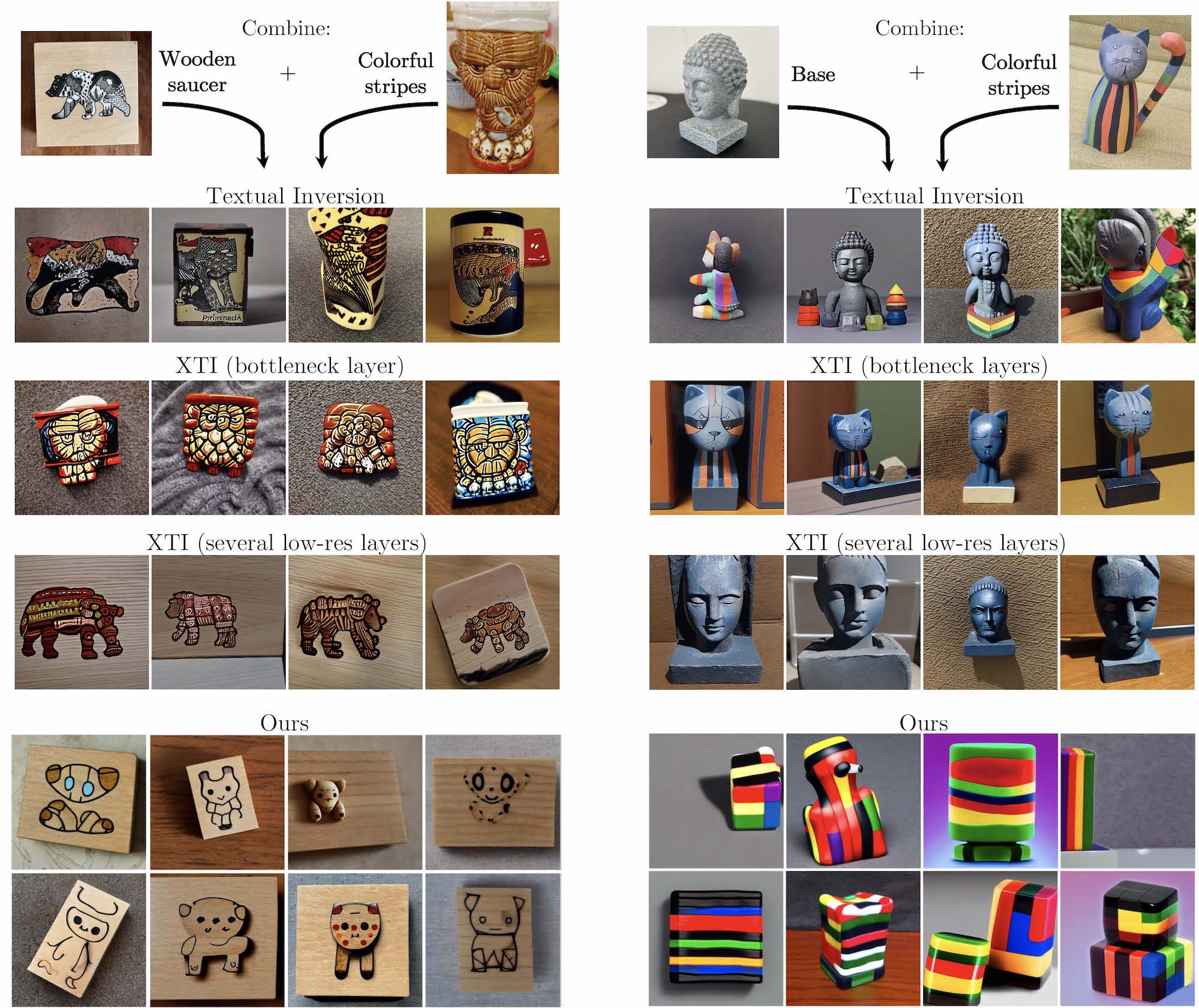}
    \caption{Comparison of blending specific aspects of concepts. Top row are the source objects and a description of which aspect is taken. Second row are the results of blending the chosen concepts using Textual Inversion \cite{gal2022textual}. Third row are the results of blending with Extended Textual Inversion \cite{Voynov2023PET} (XTI) when only bottleneck layers are provided with the left object. Fourth row are results of XTI where a wider range of the low-resolution layers are provided with the left object. Last two rows are images generated with our proposed approach.}
    \label{fig:pplus_comp}
\end{figure*}

\section{Baselines}
In the absence of existing works attempting to achieve our goal of decomposition into different aspects, we compare our performance in intra-tree combination with two existing relevant works.

We consider Textual Inversion \cite{gal2022textual}, and its more advanced modification -- Extended Textual Inversion \cite{Voynov2023PET} -- designed specifically for appearance mixing (which is most similar to our \ap{intra-tree combination}). 

We provide a qualitative comparison to these methods in \Cref{fig:pplus_comp}. On the left we aim to combine the aspect of a wooden saucer and the creature on the cup from the objects presented on top. On the right we aim to combine a part of the stone statue with some specific style aspects of the cat sculpture.

Gal et al. \shortcite{gal2022textual} propose a style transfer application, in which their method can be used to find pseudo words representing a specific style taken from a given concept, and can then be applied in combination with other concepts. To extract the style code from a given concept, they replace the training texts with prompts of the form: \ap{A painting in the style of $S^*$}.

For the TI baseline, we applied the original Textual Inversion for the first concept (from which we wish to take the structure), and for the appearance concept we used their proposed style extraction application described above. This results in a pair of textual tokens $S_1^{TI}, S_2^{TI}$ that represent each concept.
We explicitly combine these tokens in a sentence, providing the desired mixing description (e.g. \ap{$S_1^{TI}$ in the style of $S_2^{TI}$} and use it to generate an image.

Voynov et al.~\shortcite{Voynov2023PET} propose an extended textual conditioning space for a diffusion model that can be used to control style and geometry disentanglement. The main idea is to provide each diffusion UNet cross-attention layer with an independent textual prompt. The authors notice that low-resolution UNet layers are commonly responsible for geometrical attributes, while high-resolution input and output layers are responsible for style-related attributes.  

For the task of style mixing, given a pair of objects, the method performs two independent Textual Inversions to this extended prompt space (called XTI in the paper). Then, the low-resolution layers are provided with the inversion of the object that donors the shape, and the high-resolution layers are provided with the inversions of the object that donors the appearance.

For the comparison to XTI \cite{Voynov2023PET}, we use their recommended hyperparameters.
We apply two independent Textual Inversions to the extended prompt space, which brings the pair of textual tokens $S_1^{XTI}, S_2^{XTI}$. To use the geometry from $S_1^{XTI}$ and appearance of $S_2^{XTI}$, we provided the prompt \ap{a photo of $S_1^{XTI}$} to the deeper (low-res) layers and \ap{a photo of $S_2^{XTI}$} to the shallower (high-res) UNet layers.
We tried to combine the concepts using different layers split to achieve the best possible performance. 

From \Cref{fig:pplus_comp}, we can see that these baselines fail to combine the very specific aspects of the source objects. Textual Inversion commonly blends the attributes, while XTI is able to transfer either the whole creature's appearance, or texture only, failing to extract only the shape. In contrast, using our approach it is possible to pick the two distinct aspects and combine them naturally to depict a new concept.

\null
\newpage
\section{Ablation and Analysis}
\subsection{Binary Tree}
Our choice to use a binary tree stems from two main reasons: (1) complexity, and (2) consistency.

It is technically possible with our method to build a tree with more than two children per node, however, we believe this may add redundant complexity to the method.
The use of more than two children will result in a longer running time in each level since we will have to split more nodes. Additionally, after two levels we will receive 12 aspects (for a simple scenario of three nodes), which may be difficult to visualize and navigate.

\begin{figure}
    \centering
    \includegraphics[width=1\linewidth]{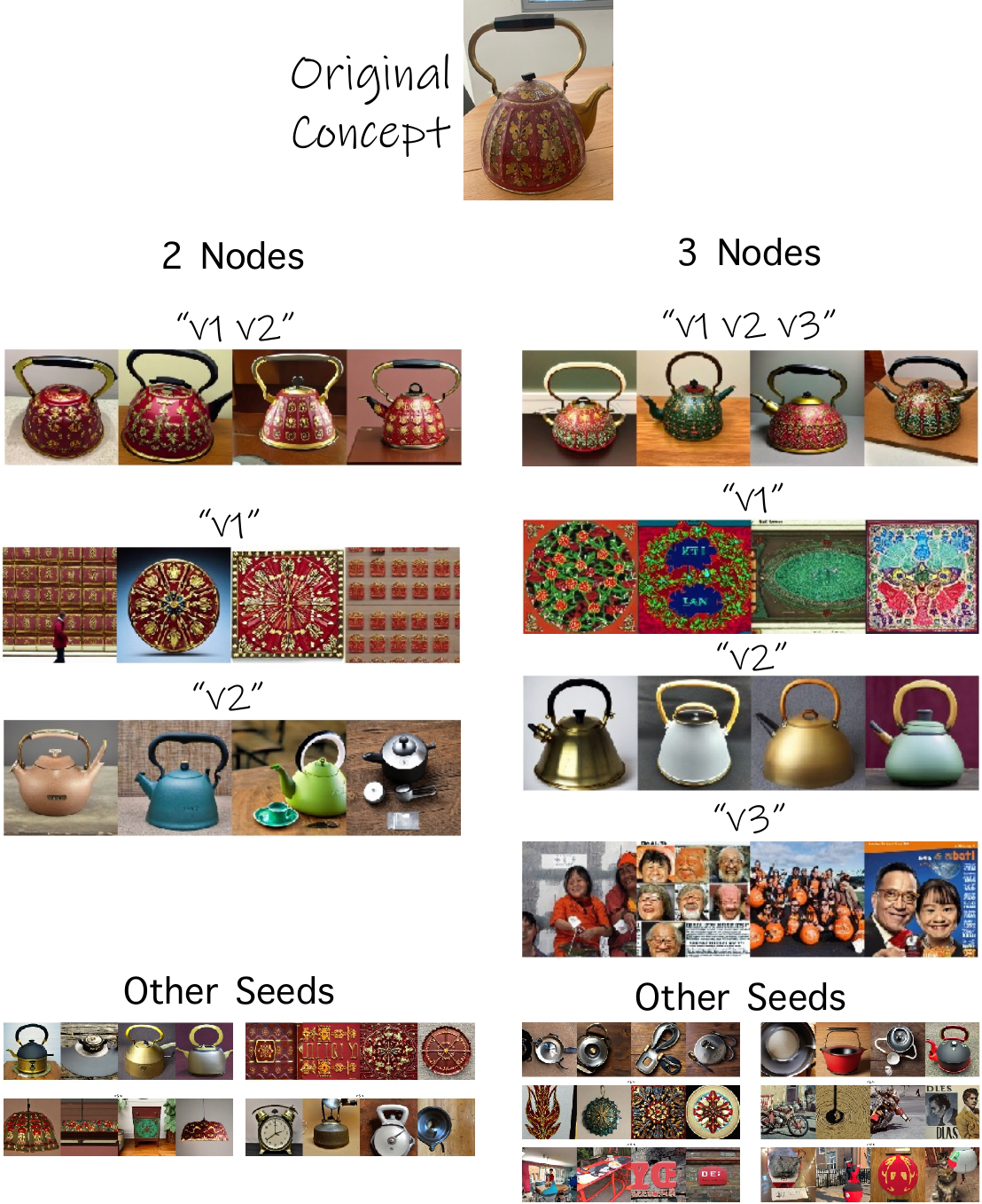}
    \caption{Comparison of optimizing for two child nodes (left) v.s. three child nodes (right). Using three nodes increases the chance of arriving at inconsistent or irrelevant concepts.
    At the top we show the results of the chosen seed among the four seeds, and below we demonstrate how two of the other seeds provide similar results, demonstrating that this trend is general.}
    \label{fig:number_of_nodes}
\end{figure}

In terms of consistency, we observe that when optimizing more than two nodes at a time, the chance of receiving inconsistent nodes increases.
Often, two nodes will be consistent, and the third node is inconsistent or may depict irrelevant concepts such as background.
We visually demonstrate this in \Cref{fig:number_of_nodes}, on the \ap{red teapot} object.
We present the aspects obtained from the optimal seed after 200 iterations, for the case of two nodes (left) and for the case of three nodes (right). As can be seen, the sub-concept in $v_3$ for the 3 nodes optimization does not appear to be consistent or comprehensible, and therefore is not useful in achieving our goal of extracting aspects from the parent concept. For the two-node case, however, the aspects obtained provide a coherent concept in addition to decomposing the object well.
At the bottom of \Cref{fig:number_of_nodes}, we show the concepts depicted by two other seeds. The results show that when using two nodes, the rest of the seeds also produce relatively consistent results, compared to the case where three nodes were used.

The following quantitative experiment further confirms this observation.
We obtained 52 trees for our set of 13 objects (using four seeds for each object as described in the main paper). Each tree is a 3-node tree with one level, resulting in a total number of 156 nodes.

\begin{figure}
    \centering
    \includegraphics[width=1\linewidth]{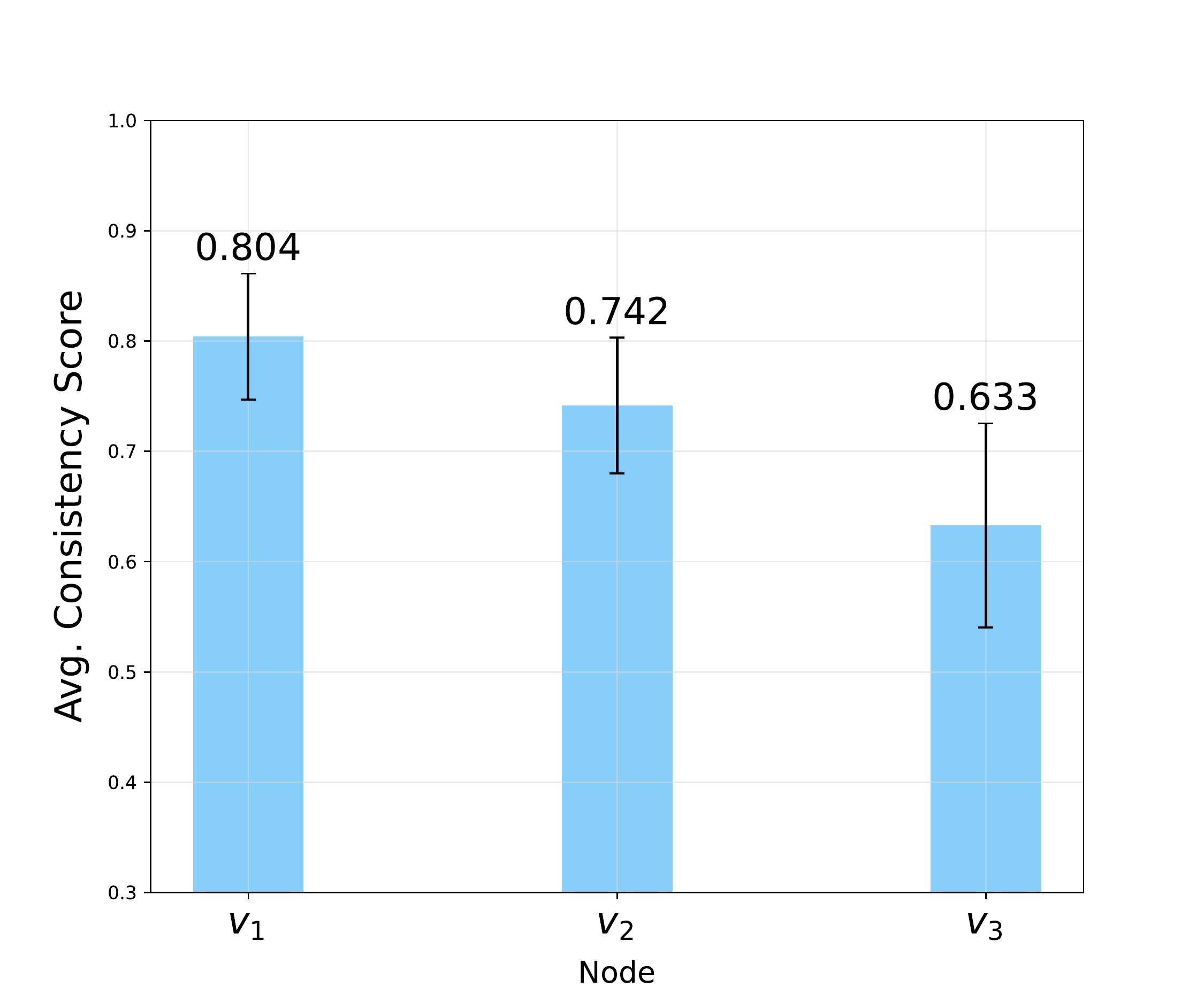}
    \vspace{-0.5cm}
    \caption{Average CLIP-based consistency scores of the three child nodes sorted from left to right. These scores were measured for 3-node trees obtained for 13 objects using 4 seeds per object.
    Observe that, on average, the third node tend to encode incoherent information, which encourages us to choose a binary tree structure.}
    \label{fig:ablation3_tokens}
\end{figure}

For each node, we then applied our CLIP-based consistency test to determine its average consistency score.
For each tree, we sorted the 3 nodes according to their consistency and received a set of $\{v_1, v_2, v_3\}$, where $v_1$ is the most consistent node of the three, $v_2$ is the second most consistent and $v_3$ is the least consistent.
We then average the scores of $\{v_1, v_2, v_3\}$ across all objects.
Results are shown in \Cref{fig:ablation3_tokens}; observe that there is a noticeable consistency gap between the top 2 nodes (achieving average scores of $0.804, 0.742$) and the third node who achieved a score of $0.633$.
This indicates that, on average, two of the three nodes are consistent, while the last may contain incoherent information.
This experiment correlates well with our visual observation (as demonstrated in \Cref{fig:number_of_nodes}).

\subsection{Timestep Sampling}
\begin{figure}
    \centering
    \includegraphics[width=1\linewidth]{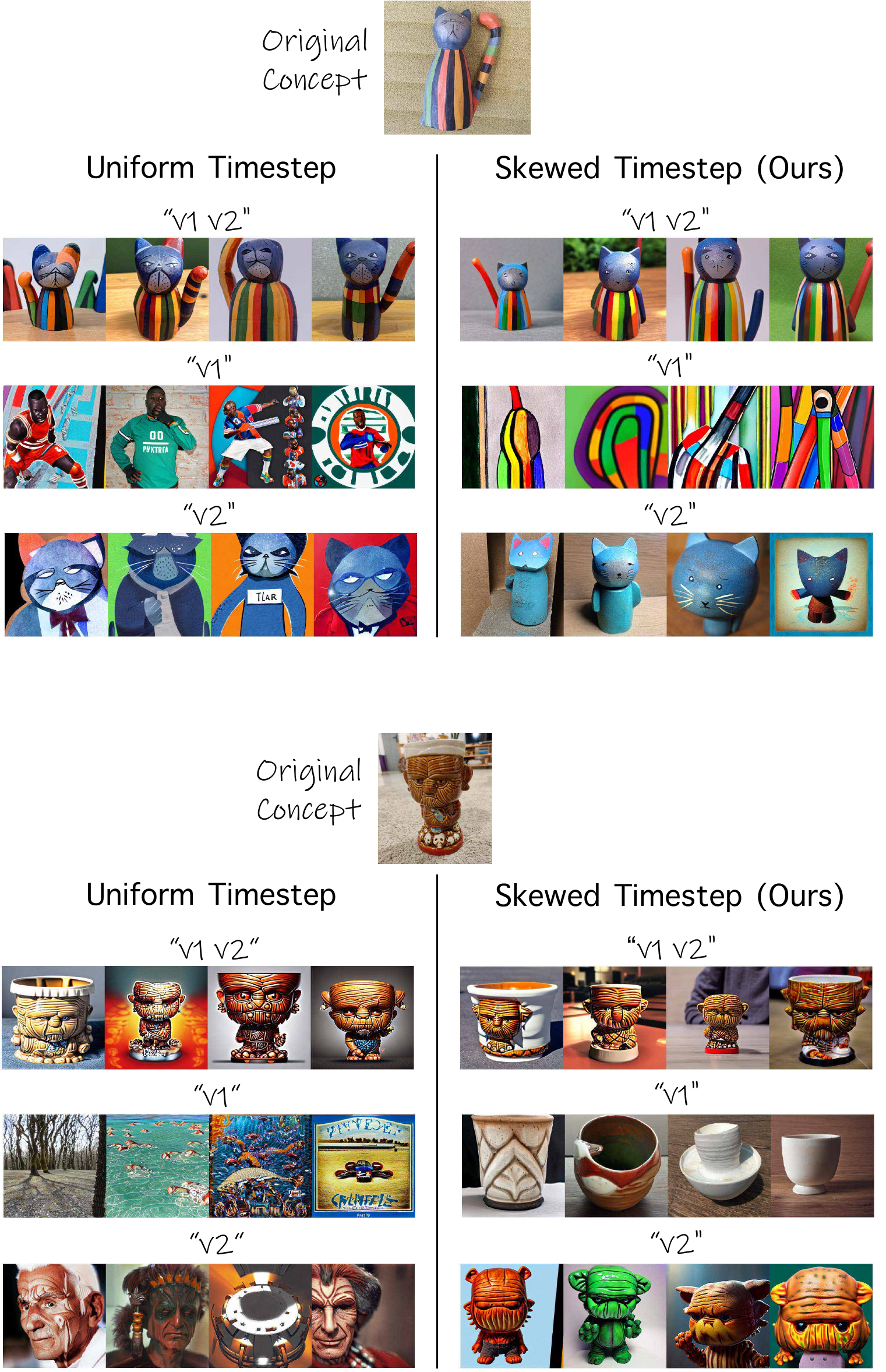}
    \caption{Timestep sampling approach ablation. We show the effect of using a uniform sampling (left), compared to using the sampling approach from ReVersion \cite{huang2023reversion}, which favor larger values of $t$.}
    \label{fig:t_sampling}
\end{figure}

As discussed in Section 4.1 of the main paper, we use the timestep sampling approach proposed in ReVersion \cite{huang2023reversion}, favoring larger $t$ values.
This sampling approach plays a significant role in the success of our method, as demonstrated in \Cref{fig:t_sampling}.

The left side of \Cref{fig:t_sampling} shows the results obtained when using a uniform sampling approach (which is the more common approach in LDM-based optimization), the right side shows the results obtained when using the sampling method we selected from ReVersion.

In both cases, the results were obtained after 500 iterations with the same seed and settings.
As can be seen, the uniform timestep sampling approach negatively affects both reconstruction quality (see \ap{v1 v2}) and decomposition quality, where for example for the cat sculpture the aspect depicted in \ap{v1} is unrelated to the original concept.

\begin{figure}[t]
    \centering
    \includegraphics[width=1\linewidth]{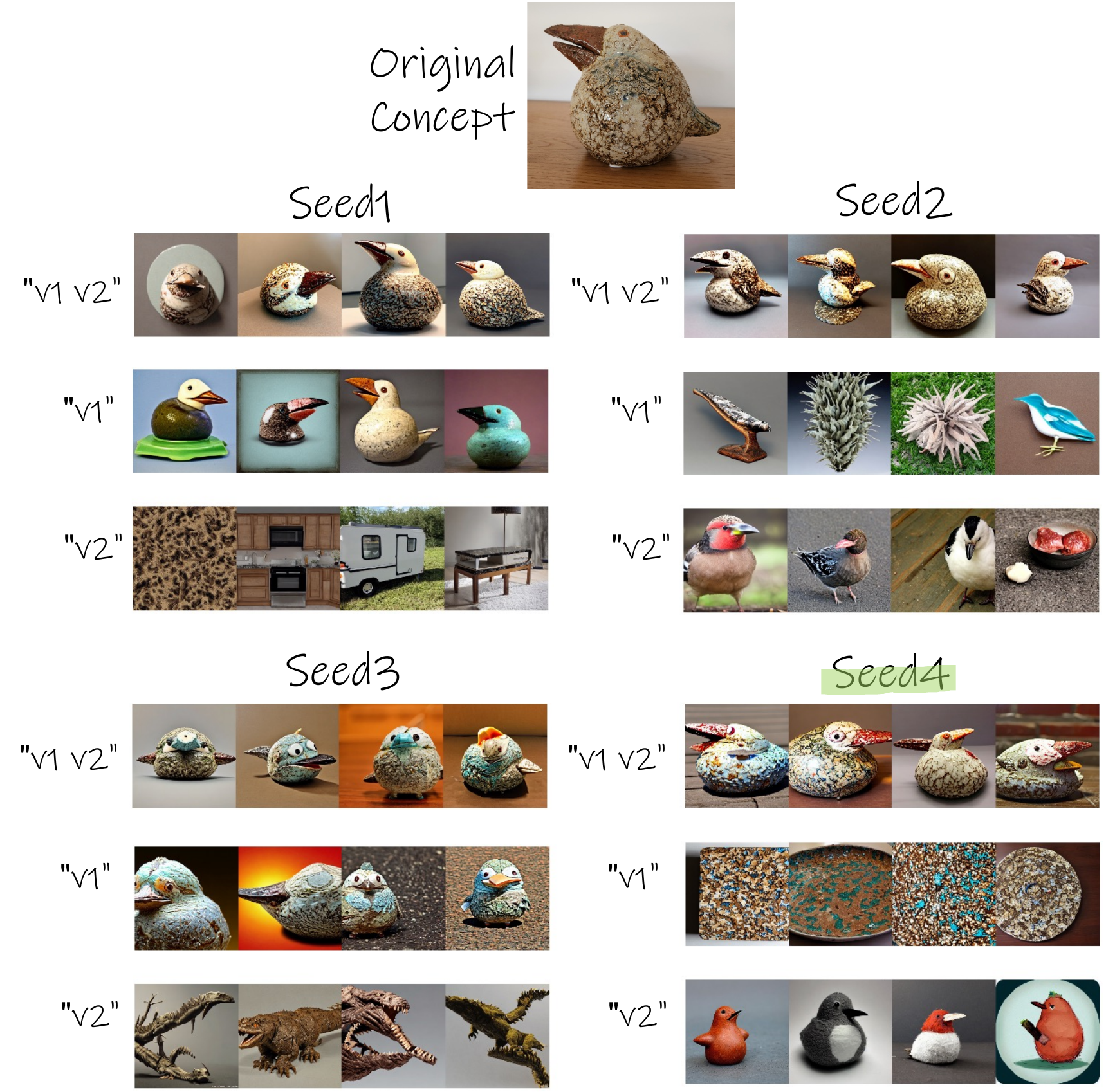}
    \caption{Results of four different seeds after 200 steps. The best seed is marked in green.}
    \label{fig:seed_bird}
\end{figure}

\begin{figure}
    \centering
    \includegraphics[width=1\linewidth]{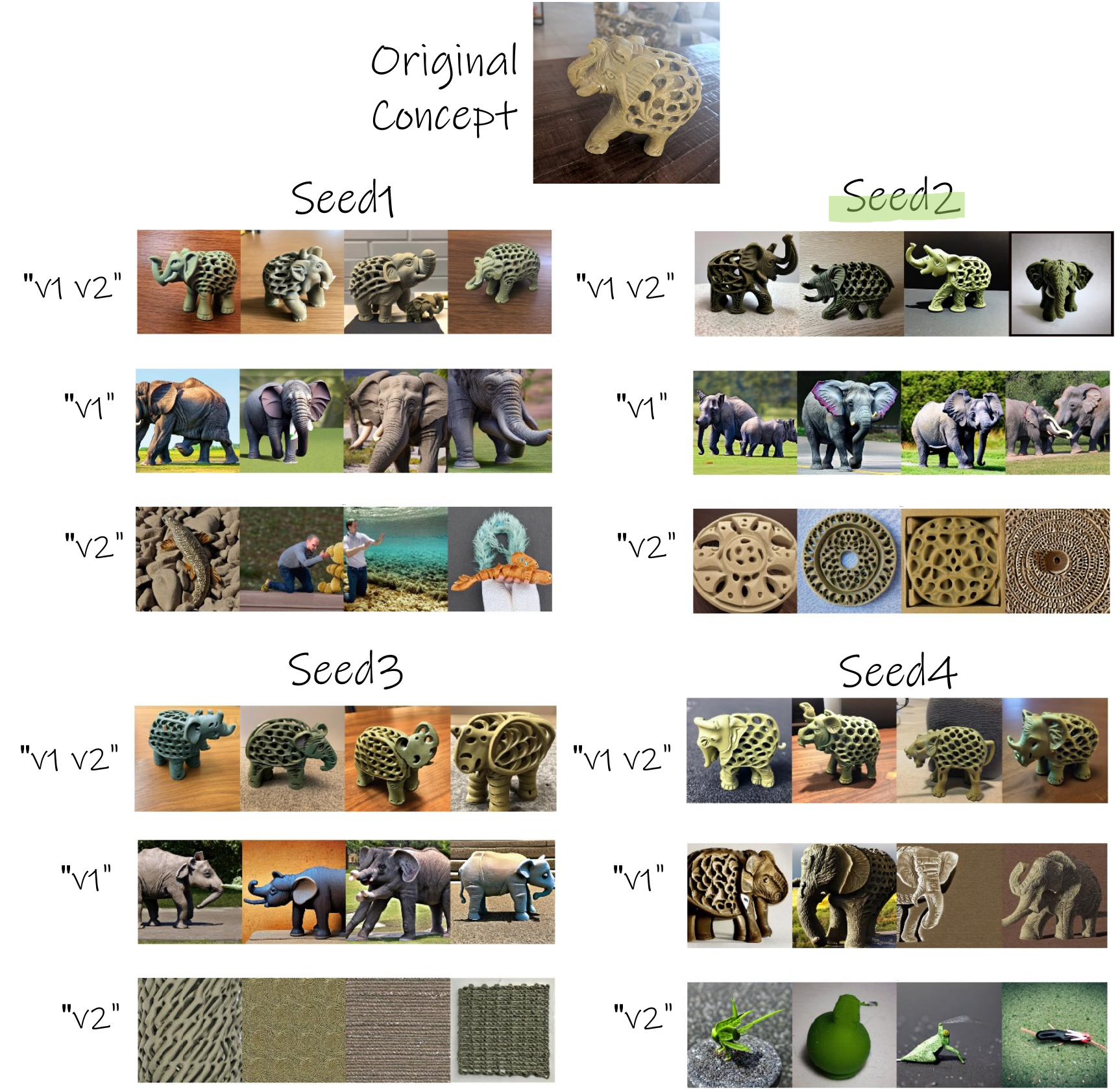}
    \caption{Results of four different seeds after 200 steps. The best seed is marked in green.}
    \label{fig:seed_elephant}
\end{figure}

\subsection{Consistency Test}
In this section we provide examples and details regarding our proposed CLIP-based consistency test presented in Section 4.1 in the main paper.
First, we visually demonstrate the effect of using $k=4$ seeds in each run.
We observe that $4$ seeds are generally enough for most of the concepts, and in most cases also $2$ seeds may be good enough.
However we do note that the variability in results among the different seeds can be quite meaningful in some cases.
We demonstrate this in \Cref{fig:seed_bird,fig:seed_elephant}, where we show the original concept on top, along with the random set of images generated for each nodes in each of the seeds.

The seed that was chosen using our CLIP-based consistency measurement is marked in green.
While the results depicted in \Cref{fig:seed_bird} were reasonable for most of the seeds, in \Cref{fig:seed_elephant} we can see that seed1 and seed2 are failure cases, where in seed 1 the concept depicted in $v_2$ is inconsistent and not interpretable, and in seed4 we have a case of one dominant node ($v_1$).

\begin{figure*}
    \centering
    \includegraphics[width=1\linewidth]{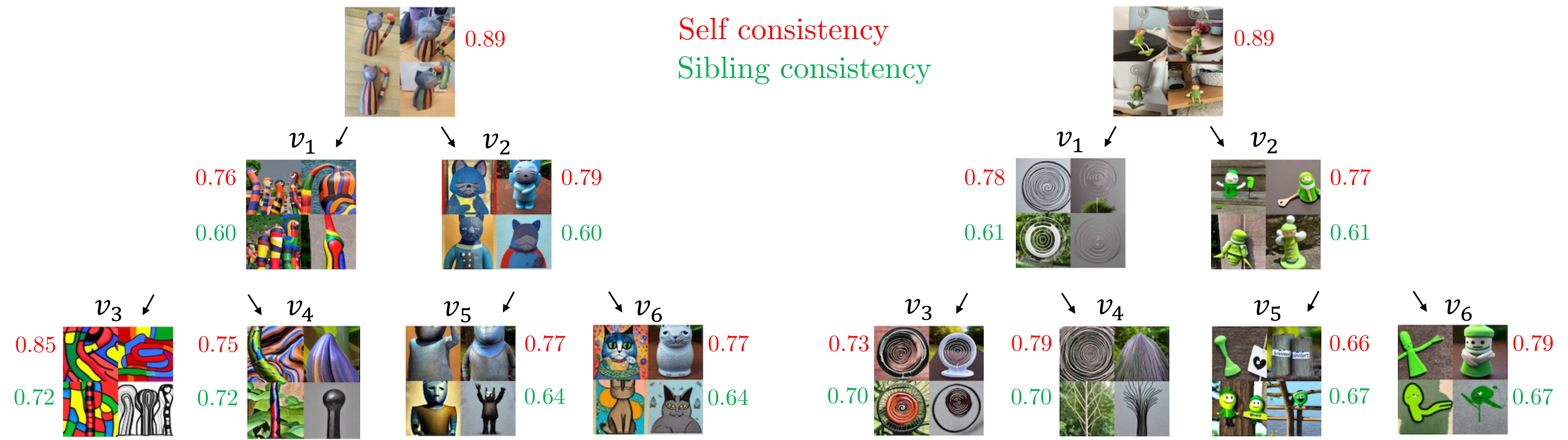}
    \caption{An illustration of two trees with different characteristics. The original training set is depicted at the root of the trees. Next to each node we present its self-consistency score (in red) and the consistency score of that node with its brother node (in green).
    The scores were obtained using our CLIP-based consistency measurement described in the main paper.}
    \vspace{-0.5cm}
    \label{fig:tree_depth}
\end{figure*}

\begin{figure}
    \centering
    \includegraphics[width=0.8\linewidth]{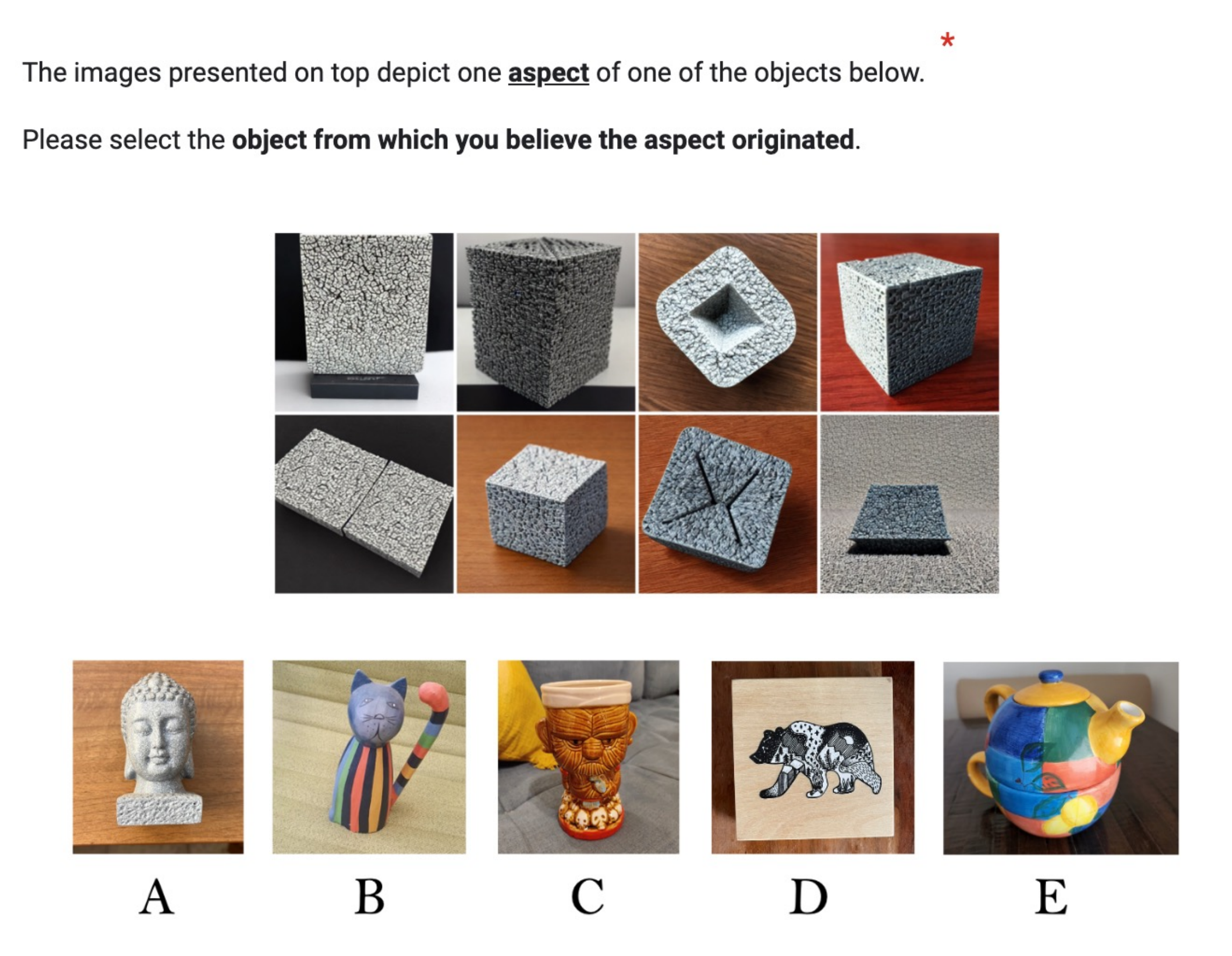}
    \caption{Example of a question we presented in the aspect relevance survey.}
    
    \label{fig:user_study_example2}
\end{figure}

Additionally, we provide an illustration to better clarify the significance of the consistency scores and their relationship to the patterns observed visually in the trees.
In \Cref{fig:tree_depth} we show examples of two trees with different characteristics.
Next to each node are the self-consistency score (marked in red) and the siblings consistency score (marked in green).
The first score measures the degree to which the images depicted in a specific node are consistent withing themselves. The second score indicates the similarity between sibling nodes.

First, observe that the self consistency score for the root node ($0.89$) is the highest, since the images depicted in that node originated from the set provided by the user. This indicates the highest consistency score possible in our settings.
In addition, we observe that the self consistency score across most nodes is relatively high and does not vary significantly as we go deeper in the tree. However, $v_5$ in the right tree obtained a self consistency score of $0.66$, which is relatively low, and in our scale it means that the set is not considered consistent.

Considering that this node is not consistent with itself, it is obvious that it is not consistent with its sibling node, which is why, in such cases, we can ignore the score obtained in green for that node in this discussion.

We now examine the scores in green, which indicate consistency across siblings.
First, note that in both trees, the consistency across siblings is low ($0.6$ and $0.61$) in the first level, suggesting that a good separation has been achieved.
However, at the second level we can see that this score generally increased, indicating that the quality of separation decreases as we go deeper in the tree.
Additionally, the sibling similarity correlates well with the visual information, with $v_3, v_4$ in the left tree and $v_3, v_4$ and $v_5, v_6$ in the right tree appearing to be more consistent than the other pairs.

It is important to note that in these cases, when the consistency among siblings is high, or when one node is inconsistent within itself, the split will be stopped at this particular level.

In order to confirm this observation, we measured these scores for the set of 13 trees that were used for the other evaluations.
For each node, we calculated the self consistency score as well as the sibling consistency score, and averaged these scores across the trees.
The results are presented in \Cref{tab:table1}.
In both levels, the average self consistency score is high, while the average siblings consistency score increased with the transition from the first to the second level, indicating that the splits are less distinct on average.
The reason for this is that as we go deeper into the tree, the components are becoming increasingly simple, making it more challenging to further split them.

\begin{table}[]
\caption{Average self consistency (left) and sibling consistency (right) scores. The scores were obtained for 13 trees.}
\begin{tabular}{ccc|ccc}

\toprule
Node & \begin{tabular}[c]{@{}c@{}}Self \\ Cons.\end{tabular} & \begin{tabular}[c]{@{}c@{}}Avg. \\ Level1\end{tabular} & Node &  \begin{tabular}[c]{@{}c@{}}Sibling \\ Cons.\end{tabular} & \begin{tabular}[c]{@{}c@{}}Avg. \\ Level2\end{tabular}                  \\ 
\midrule
v1   & 0.790            & \multicolumn{1}{c|}{\multirow{2}{*}{0.792}} & v1   & 0.580               & \multirow{2}{*}{0.58} \\
v2   & 0.794            & \multicolumn{1}{c|}{}                       & v2   & 0.580               &                       \\
\midrule
v3   & 0.781            & \multirow{4}{*}{0.783}                      & v3   & 0.711               & \multirow{4}{*}{0.69} \\
v4   & 0.780            &                                             & v4   & 0.711               &                       \\
v5   & 0.768            &                                             & v5   & 0.669               &                       \\
v6   & 0.803            &                                             & v6   & 0.669               &  \\     
\bottomrule
\end{tabular}

    \label{tab:table1}
\end{table}

\null
\newpage

\subsection{Perceptual Study}
\begin{figure}[t]
    \centering
    \includegraphics[width=0.8\linewidth]{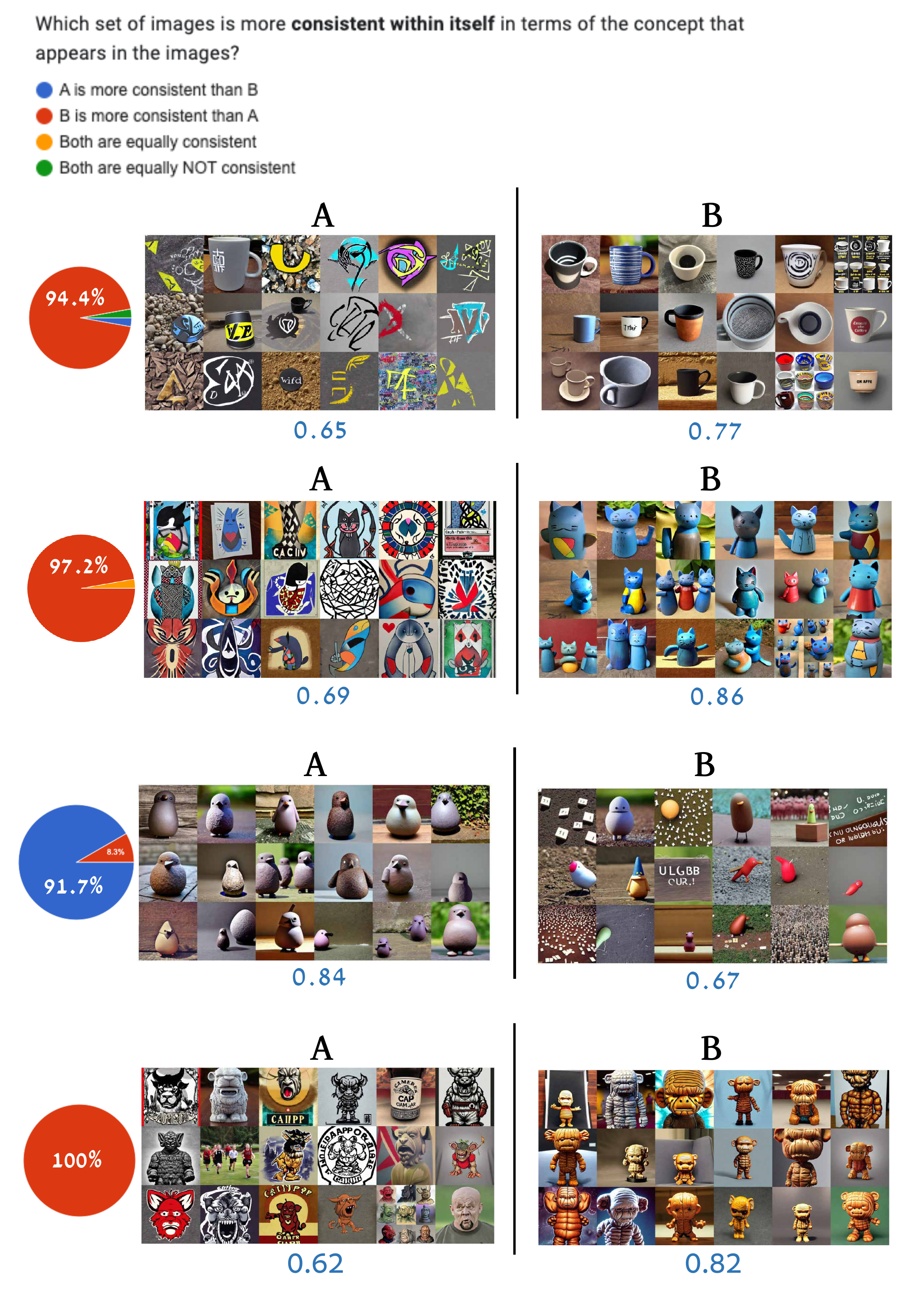}
    \caption{Examples of questions asked in the consistency evaluation survey. On the left we show the results in percentages, indicating which answer was selected by the majority of people.}
    \label{fig:consistent_user_study}
\end{figure}

The following section provides additional details regarding our perceptual study described in section 5.2 of the main paper.
For the consistency evaluation, we collected answers from 35 participants.
Participants were presented with 15 pairs of random image sets, and they were asked to determine which set in each pair is more consistent.
In order to handle cases where the sets are similar, we have also added two options to choose from - \ap{Both sets are equally consistent}, and \ap{Both sets are equally not consistent}.
\Cref{fig:consistent_user_study} contains a few examples of the survey questions.
On the left of each set, we also present the results in percentages, indicating which answer was selected by the majority of people.
In the aspect relevance experiment, we collected answers from 35 participants and asked each participant 15 questions.
\Cref{fig:user_study_example2} provides an example of the questions.
The question were obtained from 5 chosen objects, shown at the top of \Cref{fig:user_study_example2}.

\null
\newpage

\section{Additional Qualitative Results}
\begin{figure*}
    \centering
    \includegraphics[width=1\textwidth]{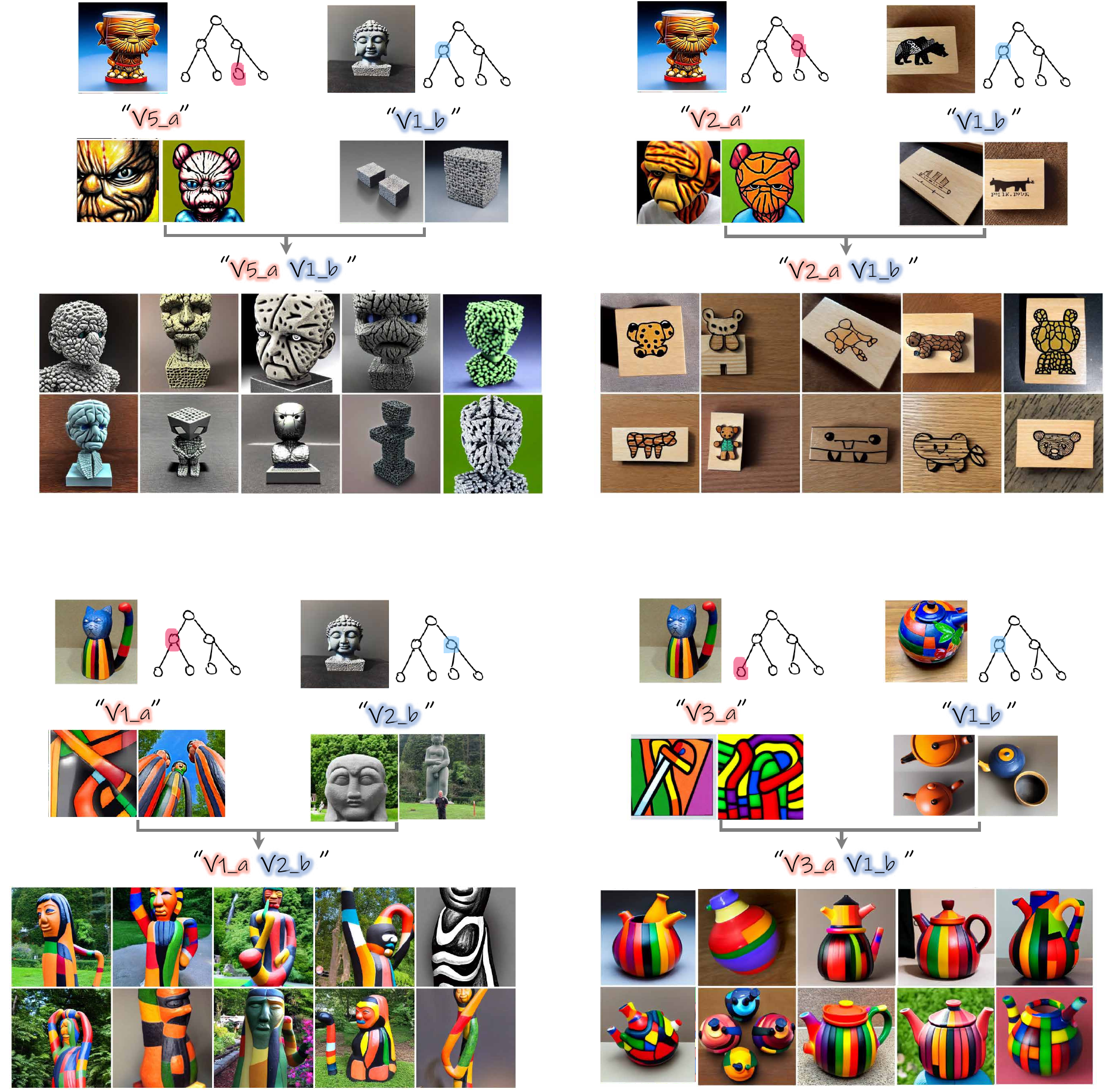}
    \caption{More examples of inter-tree combinations.}
    \label{fig:comb1}
\end{figure*}

\begin{figure*}
    \centering
    \includegraphics[width=1\textwidth]{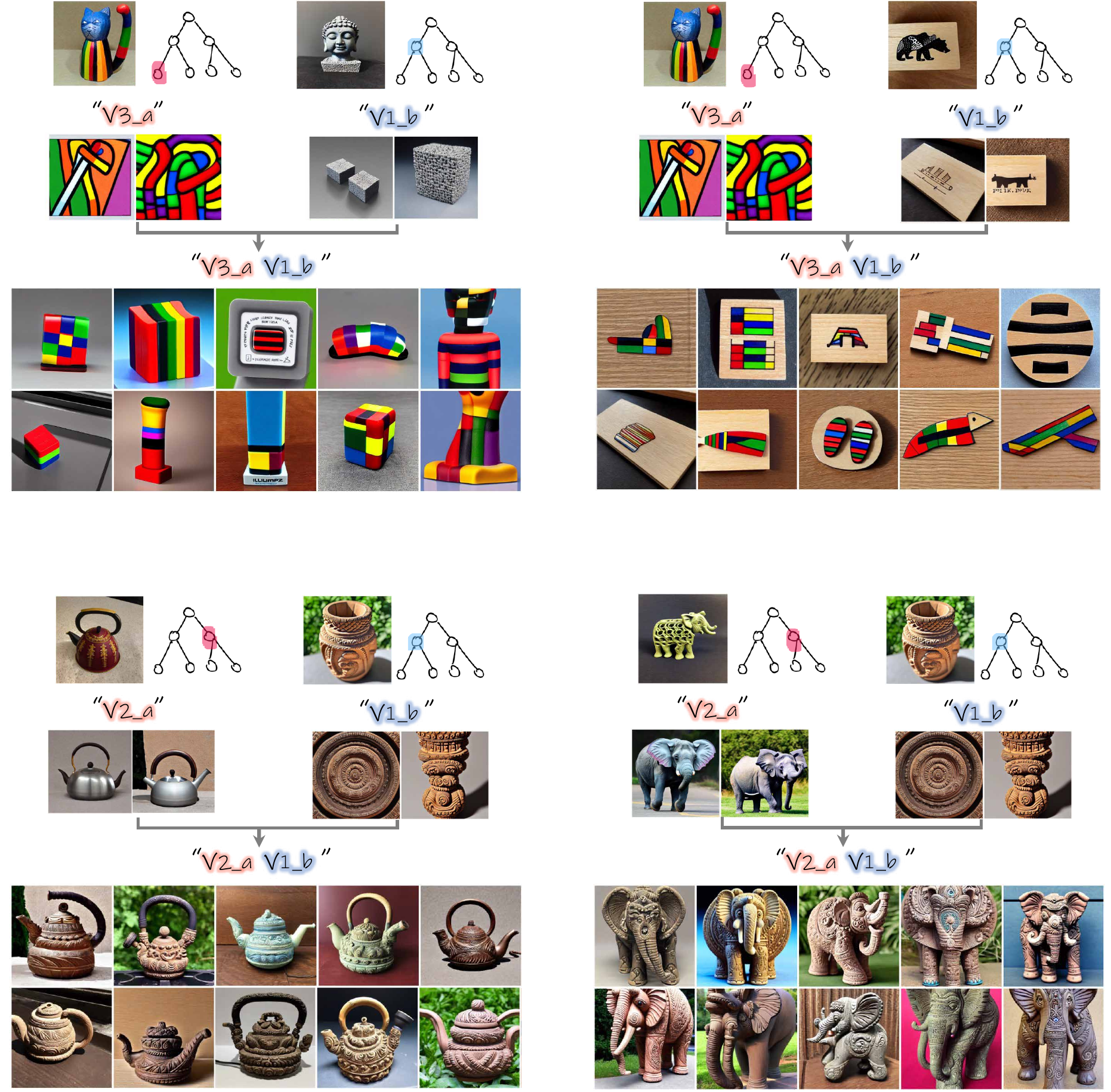}
    \caption{More examples of inter-tree combinations.}
    \label{fig:comb2}
\end{figure*}

\begin{figure*}
    \centering
    \includegraphics[width=0.9\textwidth]{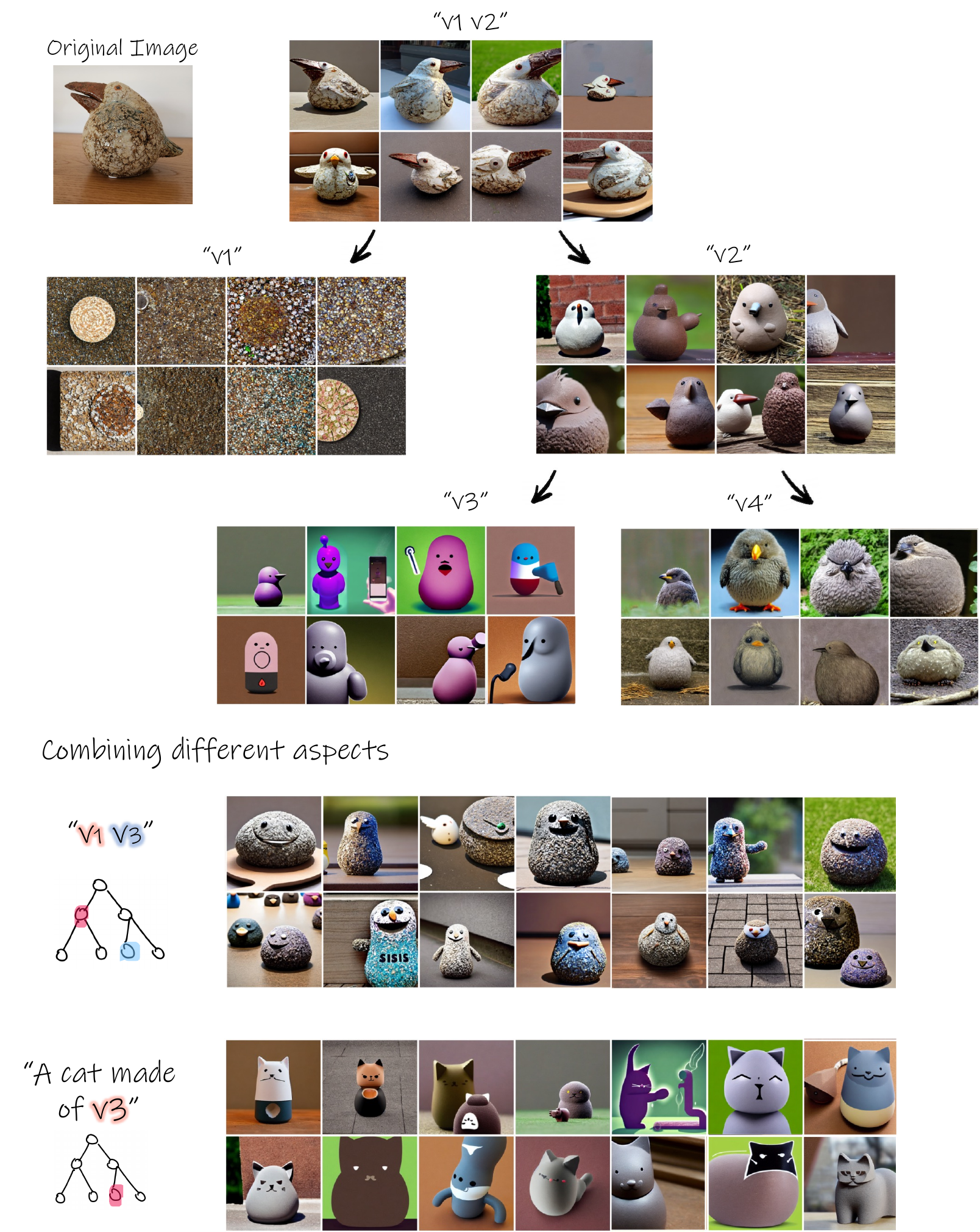}
    \caption{Exploration tree for the \ap{round bird} object. At the bottom we show examples of possible intra-tree combinations and text-based generation.}
    \label{fig:round_bird}
\end{figure*}

\begin{figure*}
    \centering
    \includegraphics[width=0.9\textwidth]{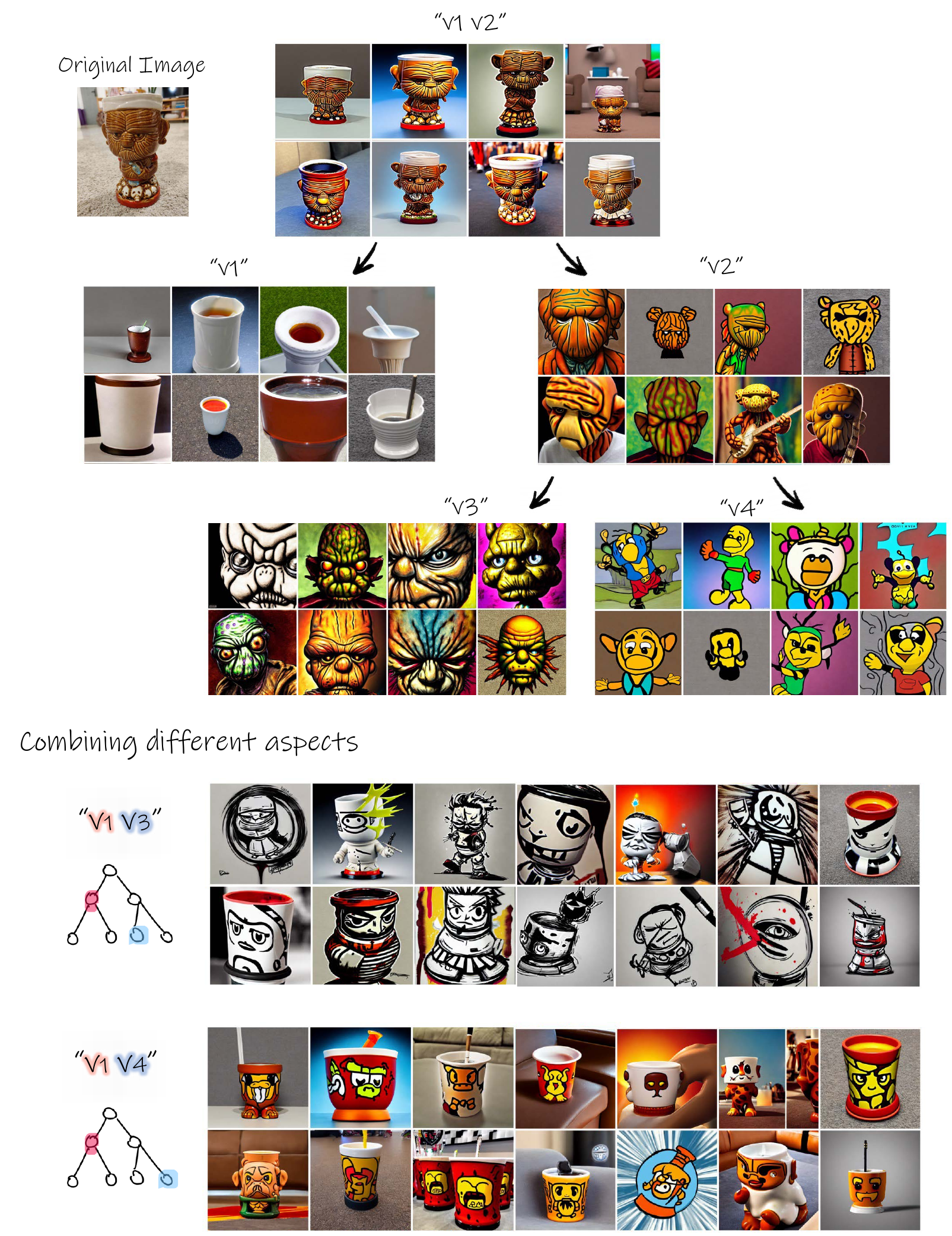}
    \caption{Exploration tree for the \ap{scary mug} object. At the bottom we show examples of possible intra-tree combinations.}
    \label{fig:mug_tree}
\end{figure*}

\begin{figure*}
    \centering
    \includegraphics[width=0.95\textwidth]{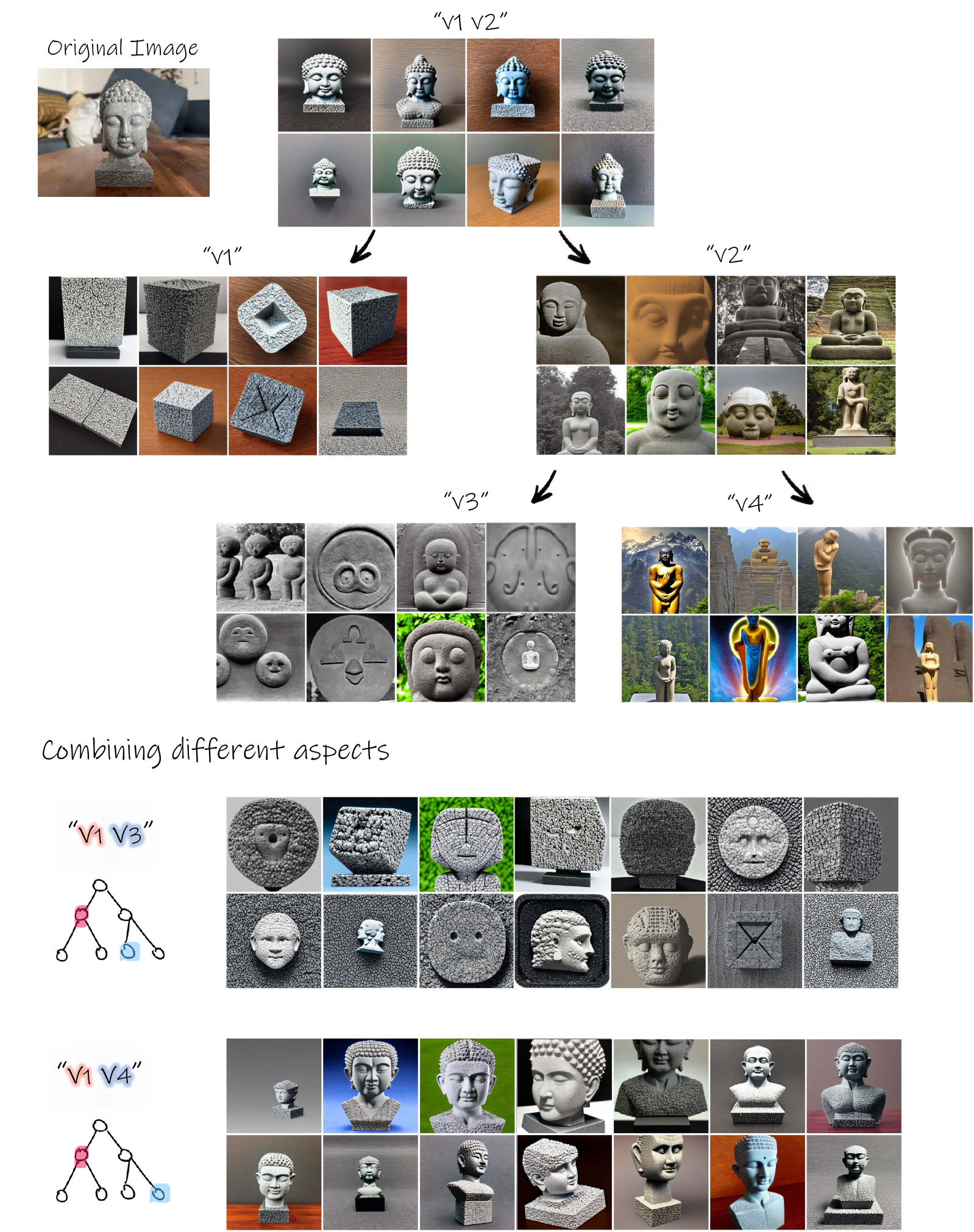}
    \caption{Exploration tree for the \ap{Buddha sculpture} object. At the bottom we show examples of possible intra-tree combinations.}
    \label{fig:buddha_tree}
\end{figure*}

\begin{figure*}
    \centering
    \includegraphics[width=0.9\textwidth]{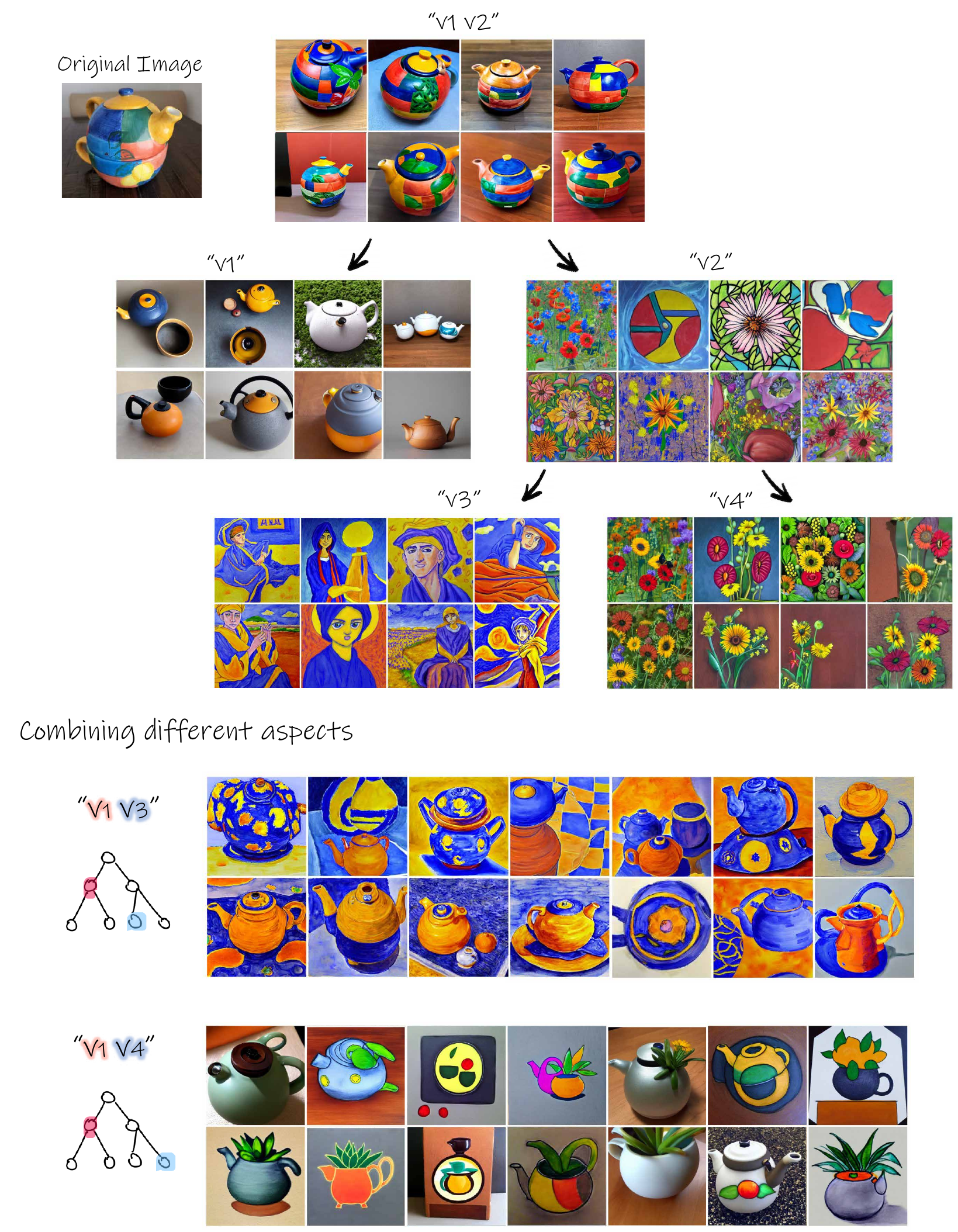}
    \caption{Exploration tree for the \ap{colorful teapot} object. At the bottom we show examples of possible intra-tree combinations.}
    \label{fig:teapot_tree}
\end{figure*}

\begin{figure*}
    \centering
    \includegraphics[width=0.83\textwidth]{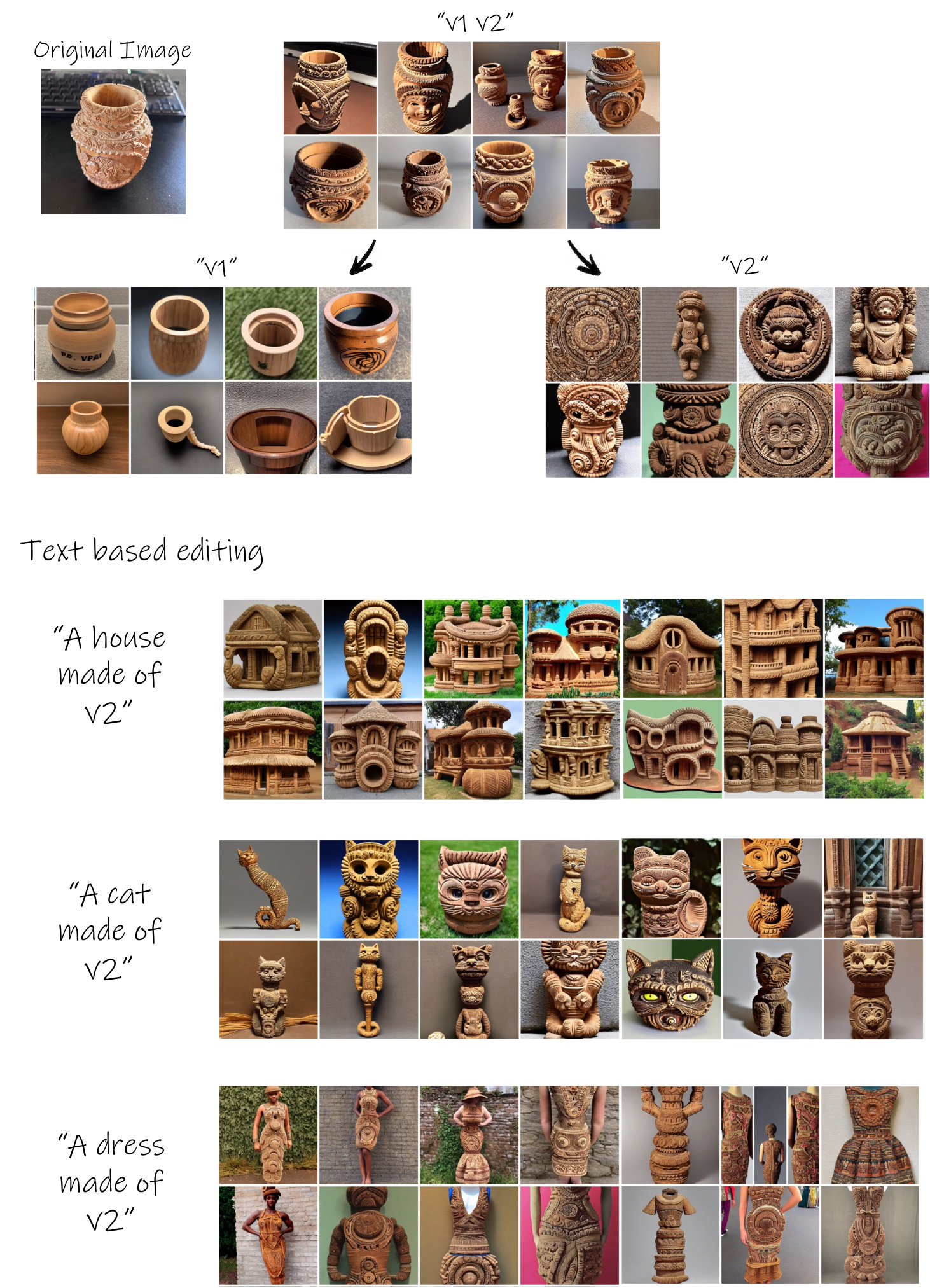}
    \caption{Exploration tree for the \ap{wooden pot} object. At the bottom we show examples of possible text-based generation.}
    \label{fig:wooden_pot}
\end{figure*}

\begin{figure*}
    \centering
    \includegraphics[width=0.83\textwidth]{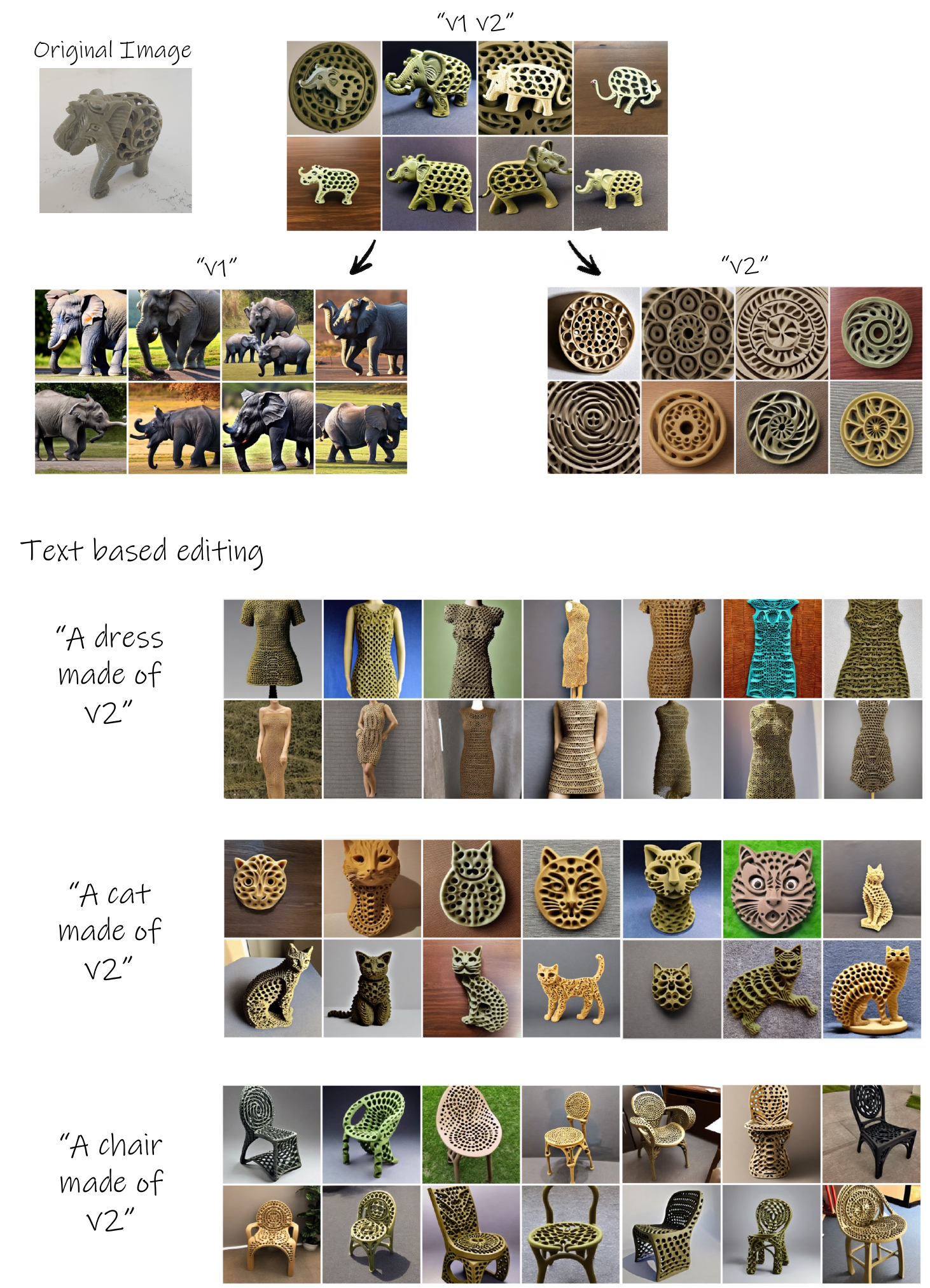}
    \caption{Exploration tree for the \ap{elephant} object. At the bottom we show examples of possible text-based generation.}
    \label{fig:tree_elephant}
\end{figure*}

\begin{figure*}
    \centering
    \includegraphics[width=1\textwidth]{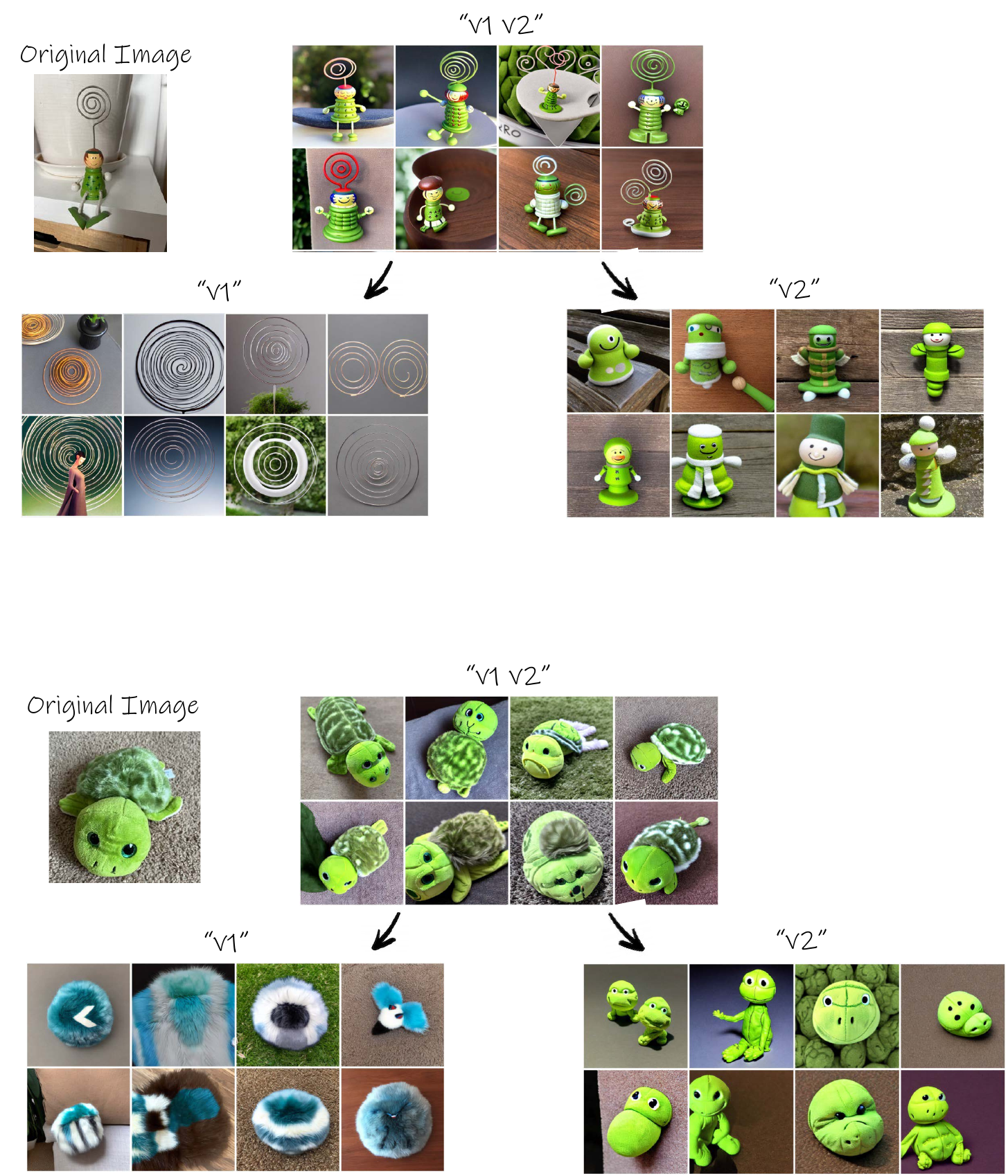}
    \caption{Exploration trees for the \ap{green doll} and the \ap{turtle} objects.}
    \label{fig:green_dall_turtle}
\end{figure*}

\begin{figure*}
    \centering
    \includegraphics[width=1\textwidth]{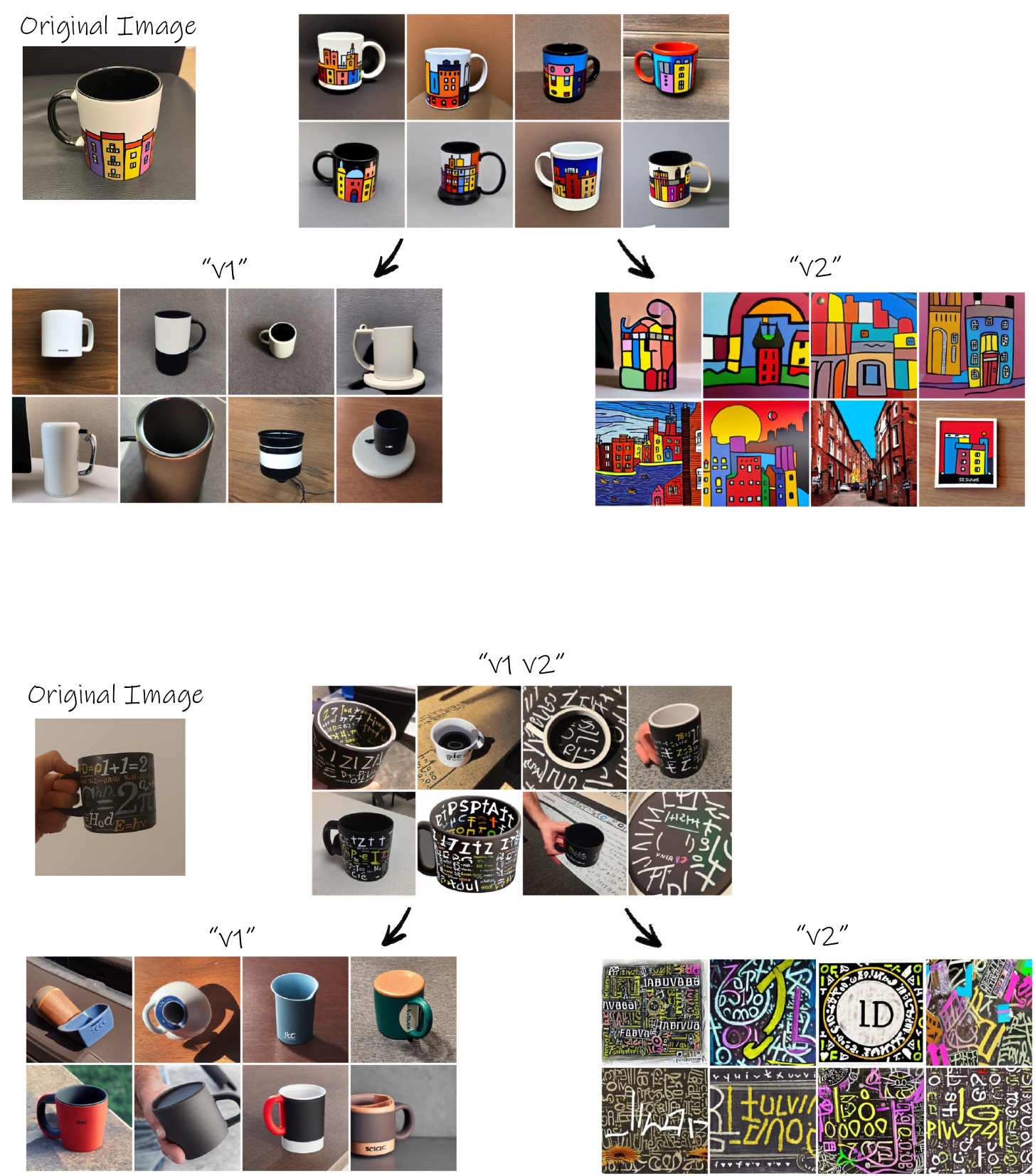}
    \caption{Exploration trees for the \ap{Girona mug} and the \ap{physics mug}.}
    \label{fig:trees_mugs}
\end{figure*}

In \Cref{fig:comb1,fig:comb2} we show more examples of inter-tree combinations.
At the top part of \Cref{fig:mug_tree,fig:buddha_tree,fig:teapot_tree,fig:round_bird,fig:wooden_pot,fig:tree_elephant,fig:green_dall_turtle,fig:trees_mugs} we show examples of trees on various objects.

\begin{figure*}
    \centering
    \includegraphics[width=0.9\textwidth]{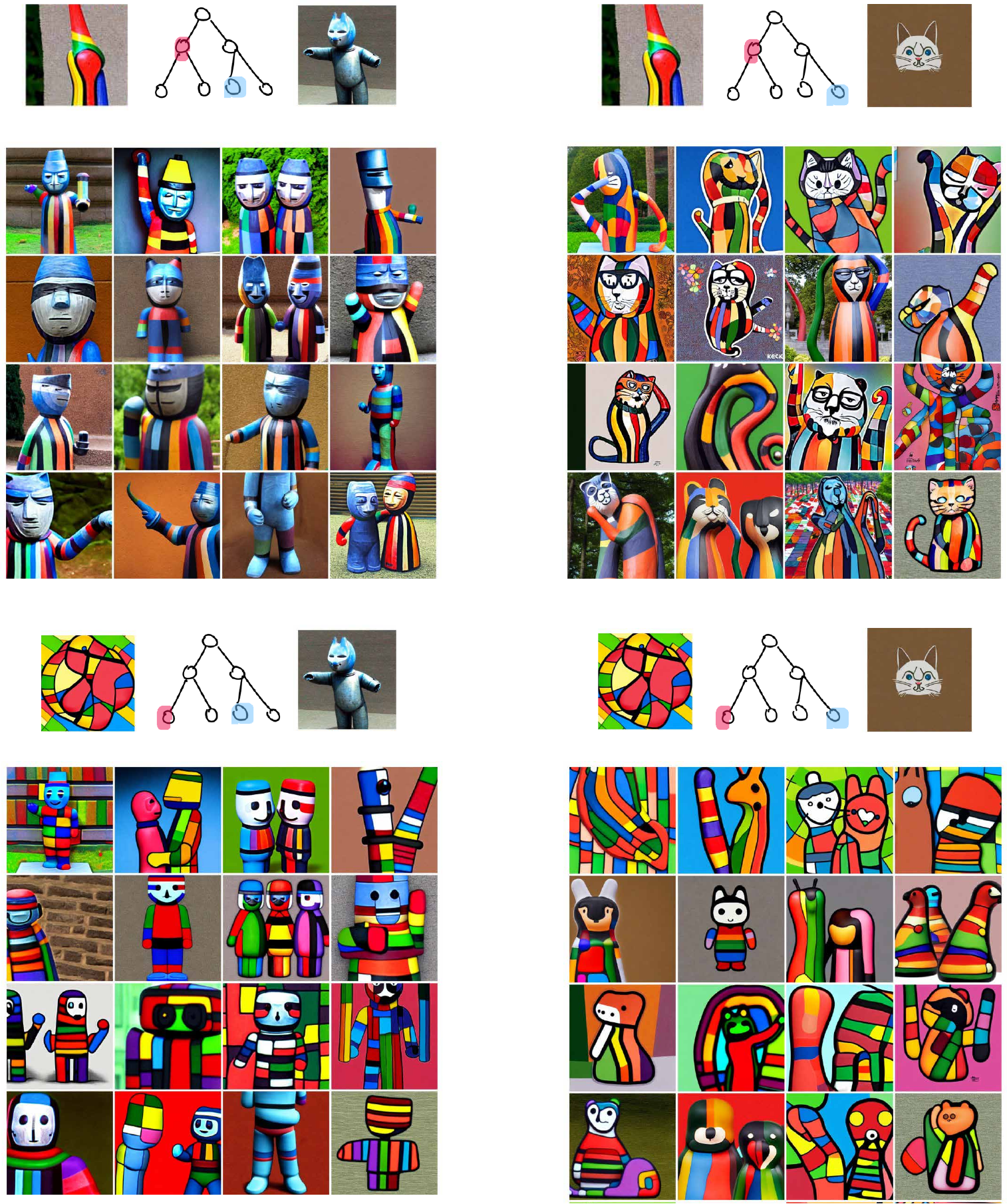}
    \caption{More examples of intra-tree combinations for the \ap{cat sculpture} object. The full original tree is shown in the main paper.}
    \label{fig:cat_intra_edit1}
\end{figure*}

\begin{figure*}
    \centering
    \includegraphics[width=0.9\textwidth]{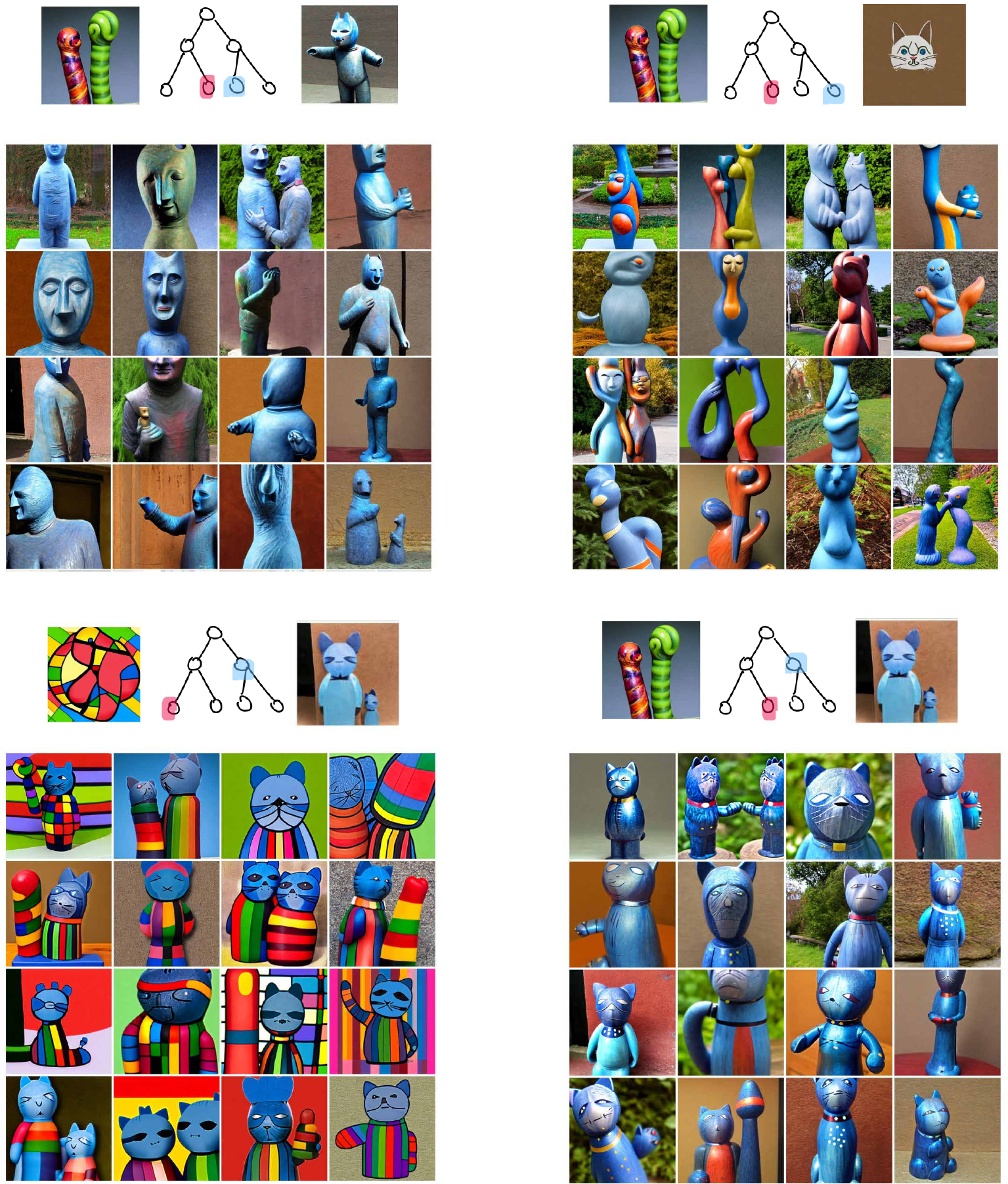}
    \caption{More examples of intra-tree combinations for the \ap{cat sculpture} object. The full original tree is shown in the main paper.}
    \label{fig:cat_intra_edit2}
\end{figure*}

At the bottom part of \Cref{fig:mug_tree,fig:buddha_tree,fig:teapot_tree,fig:round_bird} and in \Cref{fig:cat_intra_edit1,fig:cat_intra_edit2} we show visual examples of intra tree combinations.

\begin{figure*}
    \centering
    \includegraphics[width=0.9\textwidth]{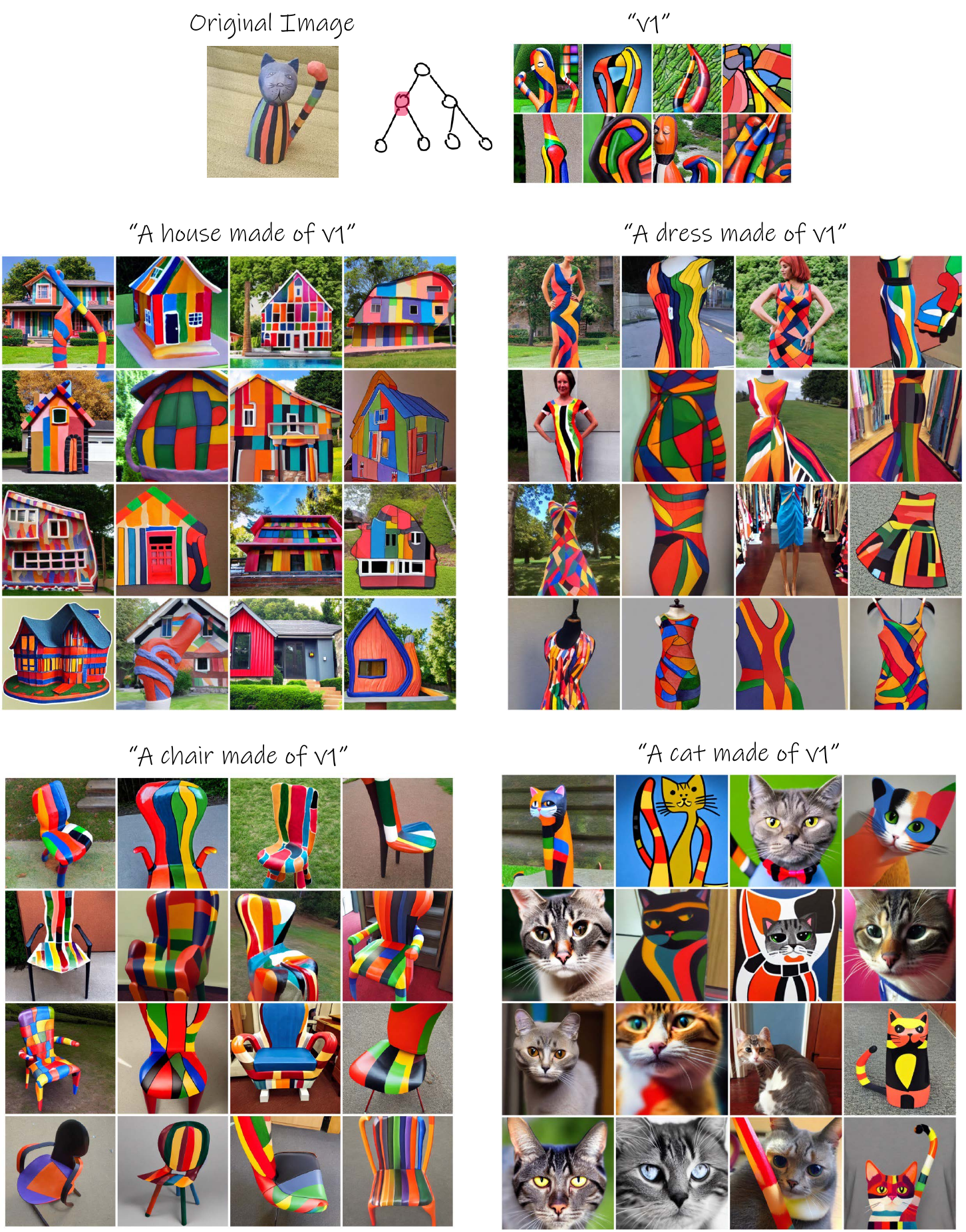}
    \caption{More examples of text based generation for the \ap{cat sculpture} object. The full original tree is shown in the main paper.}
    \label{fig:cat_text_editing1}
\end{figure*}

\begin{figure*}
    \centering
    \includegraphics[width=0.9\textwidth]{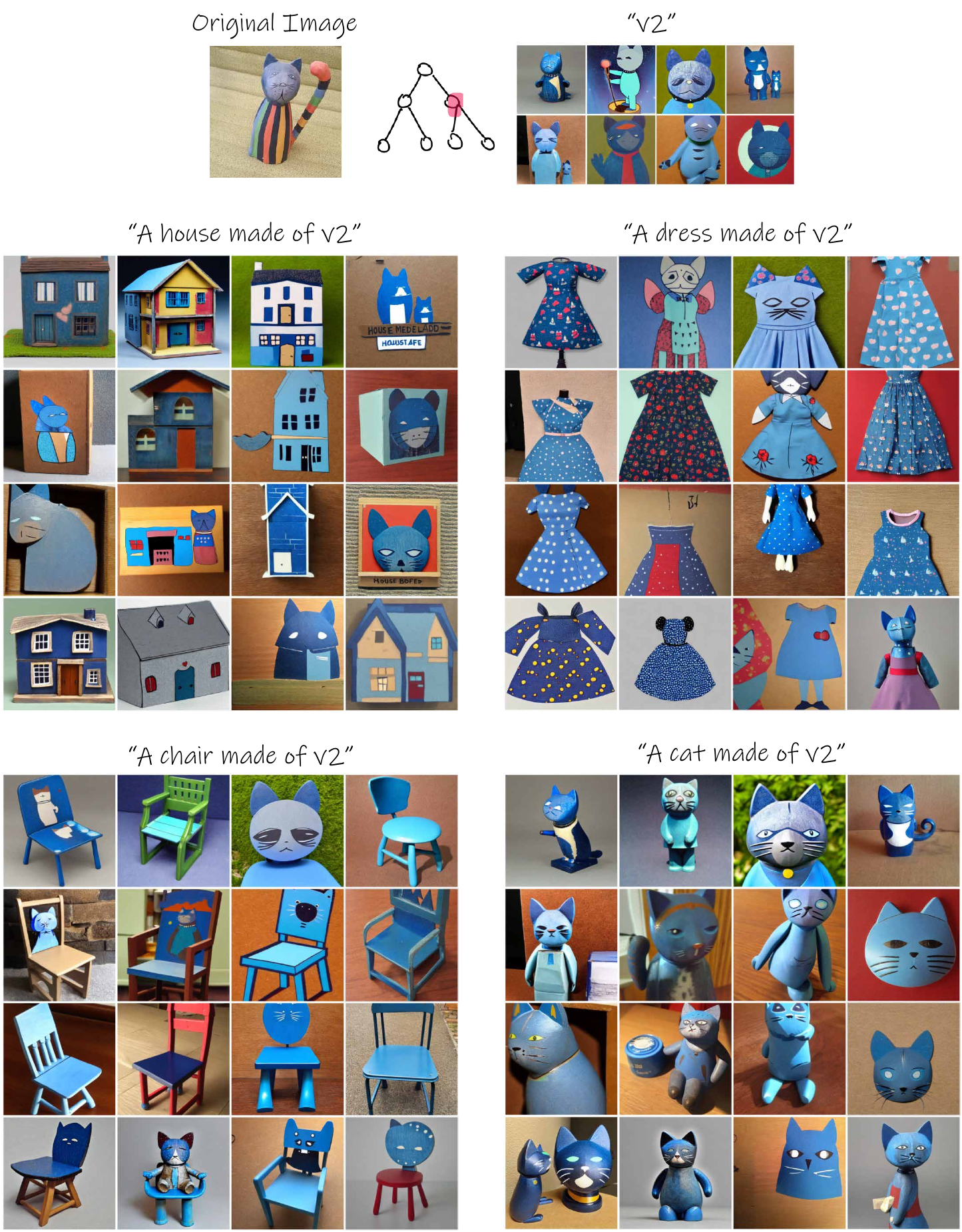}
    \caption{More examples of text based generation for the \ap{cat sculpture} object. The full original tree is shown in the main paper.}
    \label{fig:cat_text_editing2}
\end{figure*}

\begin{figure*}
    \centering
    \includegraphics[width=0.9\textwidth]{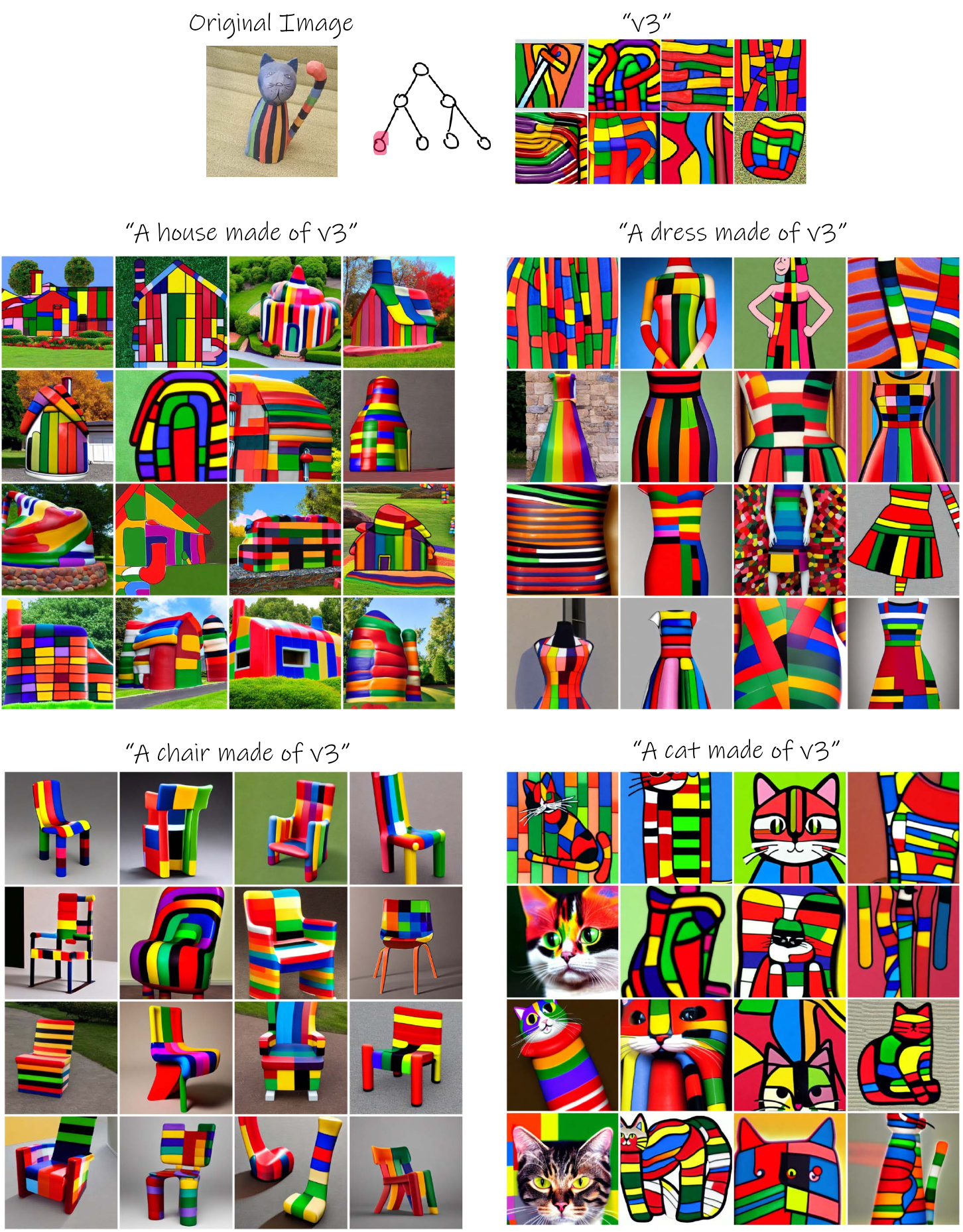}
    \caption{More examples of text based generation for the \ap{cat sculpture} object. The full original tree is shown in the main paper.}
    \label{fig:cat_text_editing3}
\end{figure*}

\begin{figure*}
    \centering
    \includegraphics[width=0.9\textwidth]{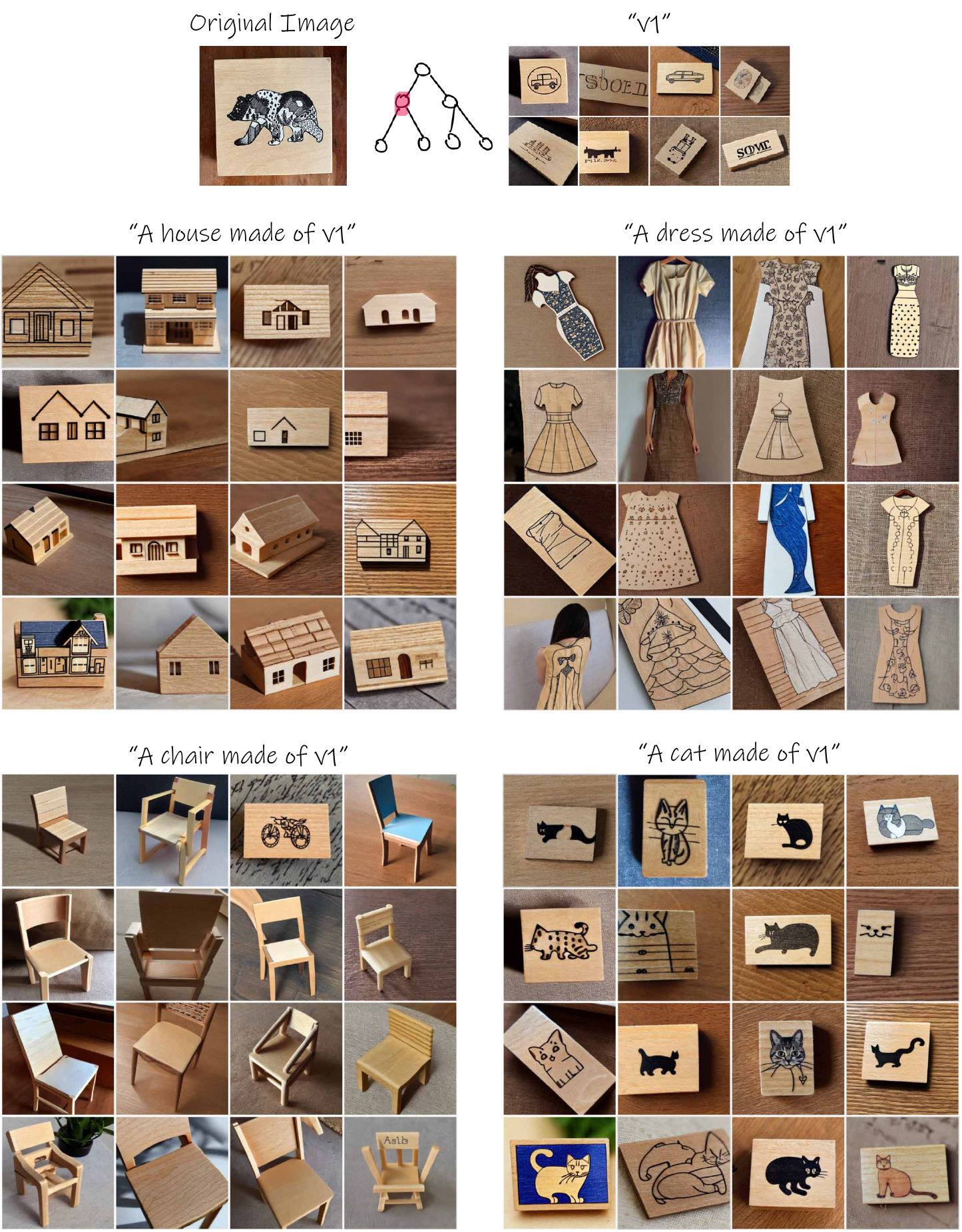}
    \caption{More examples of text based generation for the \ap{wooden saucer bear} object. The full original tree is shown in the main paper.}
    \label{fig:bear_text_editing1}
\end{figure*}

\begin{figure*}
    \centering
    \includegraphics[width=0.9\textwidth]{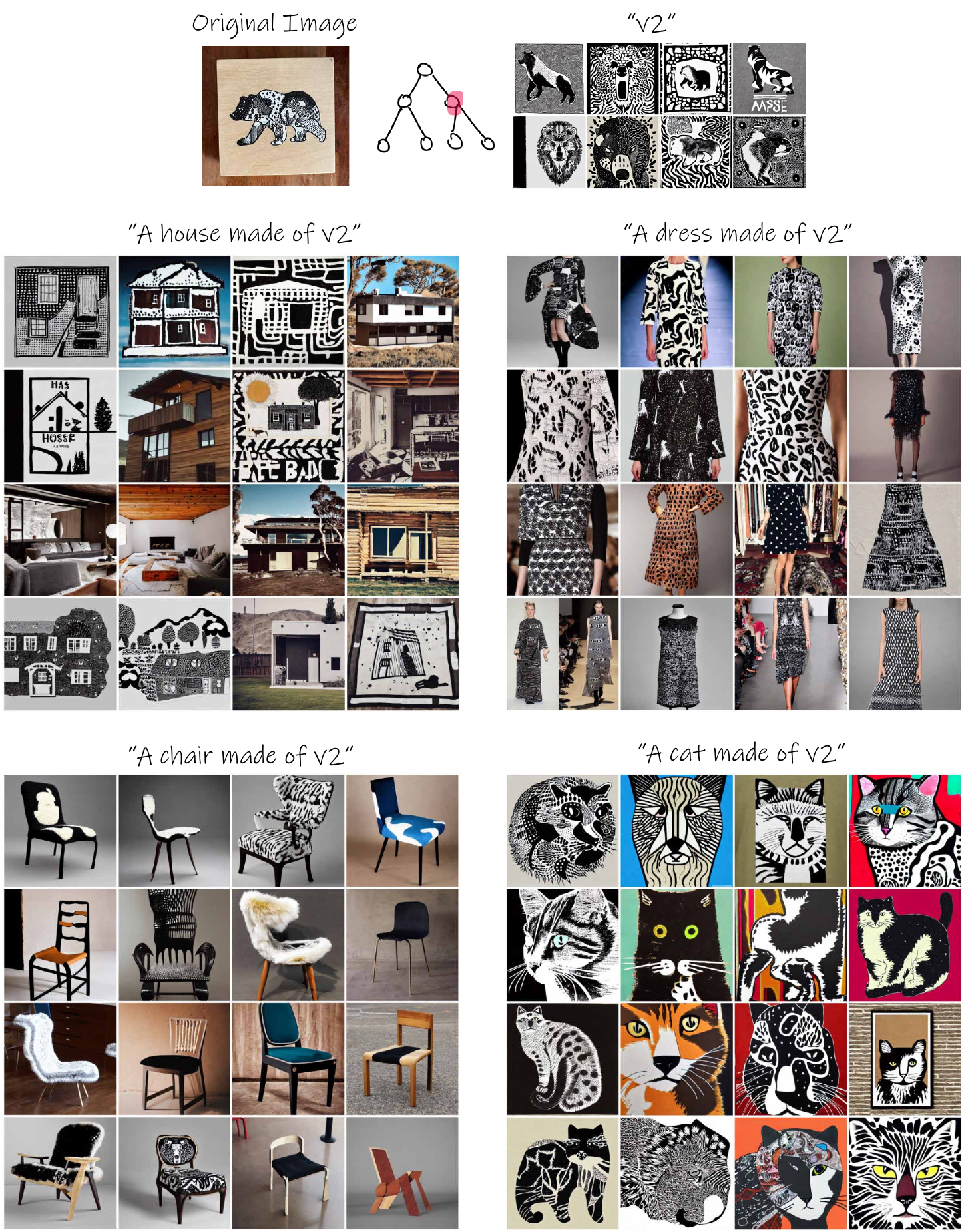}
    \caption{More examples of text based generation for the \ap{wooden saucer bear} object. The full original tree is shown in the main paper.}
    \label{fig:bear_text_editing2}
\end{figure*}

\begin{figure*}
    \centering
    \includegraphics[width=0.9\textwidth]{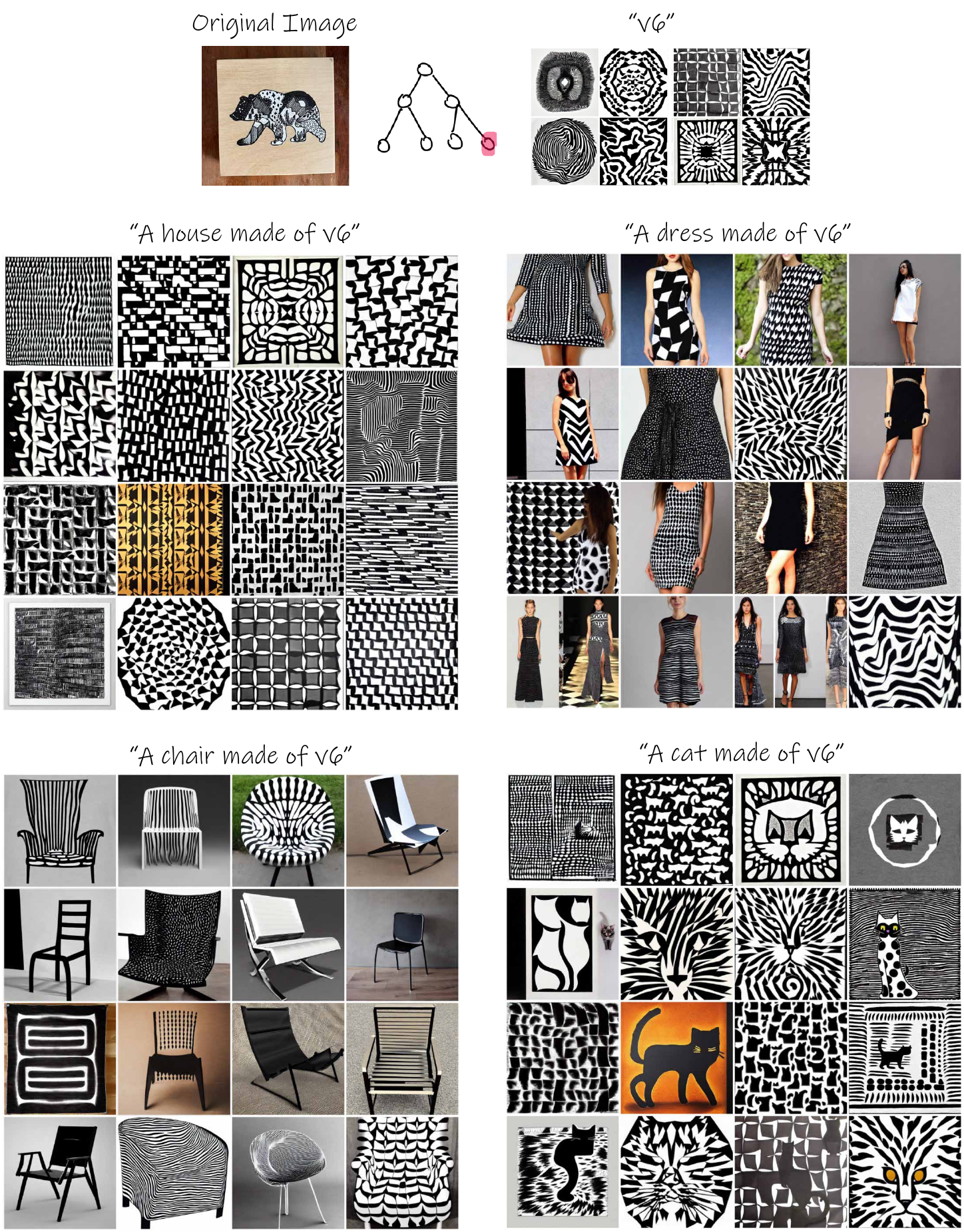}
    \caption{More examples of text based generation for the \ap{wooden saucer bear} object. The full original tree is shown in the main paper.}
    \label{fig:bear_text_editing3}
\end{figure*}

\begin{figure*}
    \centering
    \includegraphics[width=0.9\textwidth]{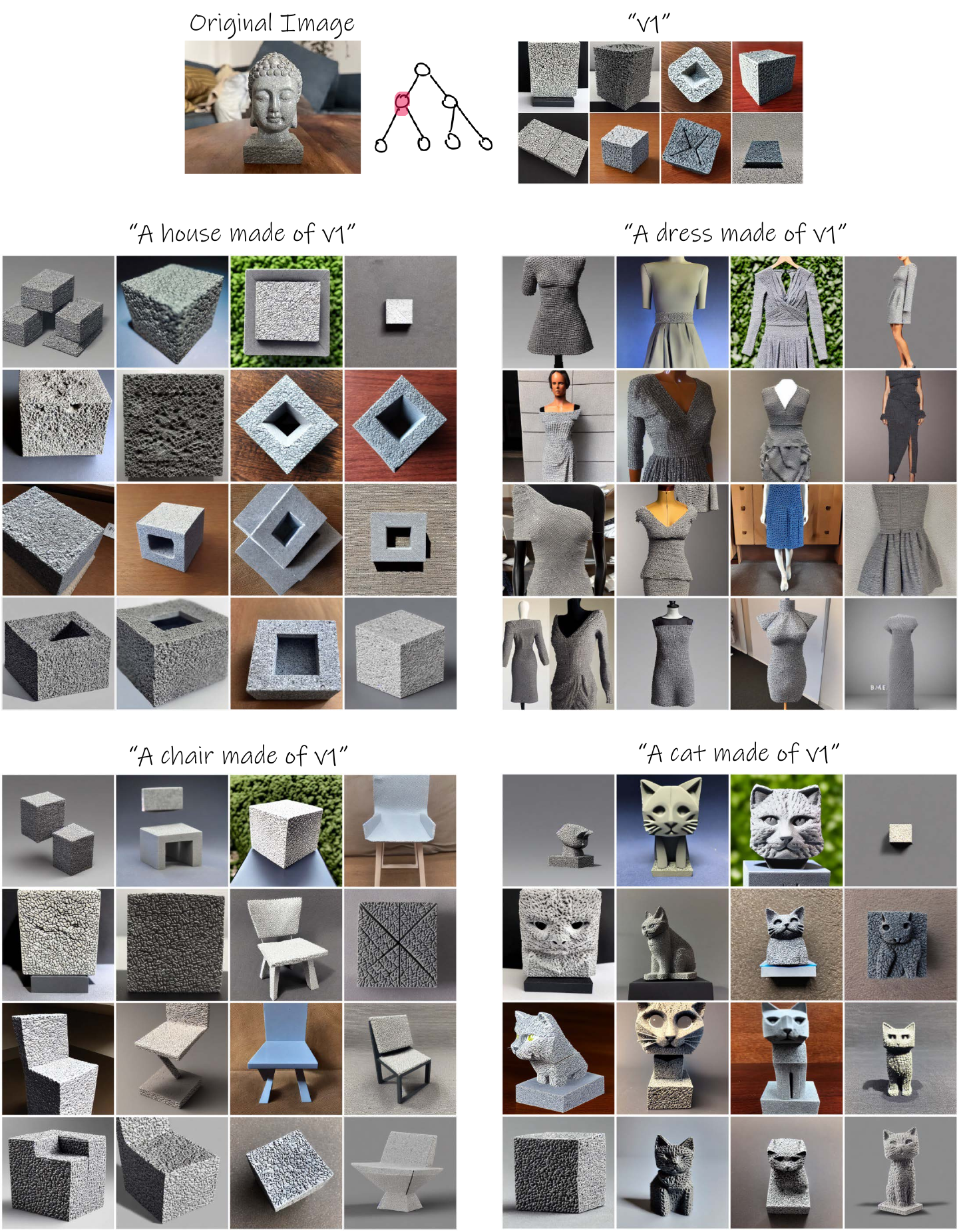}
    \caption{More examples of text based generation for the \ap{Buddha sculpture} object.}
    \label{fig:buddha_text_editing1}
\end{figure*}

At the bottom part of \Cref{fig:round_bird,fig:wooden_pot,fig:tree_elephant} and in \Cref{fig:cat_text_editing1,fig:cat_text_editing2,fig:cat_text_editing3,fig:bear_text_editing1,fig:bear_text_editing2,fig:bear_text_editing3,,fig:buddha_text_editing1} we show examples of text based generation.

\null
\newpage
\null
\newpage
\null
\newpage
\null
\newpage
\null
\newpage
\null
\newpage
\end{document}